\documentclass[journal,compsoc,twoside]{IEEEtran}

\ifCLASSOPTIONcompsoc
  \usepackage[nocompress]{cite}
\else
  \usepackage{cite}
\fi

\usepackage{graphicx}
\usepackage{amsmath}
\usepackage{mathrsfs}
\usepackage{array}

\makeatletter		

\newcommand{\Rmnum}[1]{\expandafter\@slowromancap\romannumeral #1@}
\makeatother

\usepackage{algorithm}
\usepackage{algorithmic}
\usepackage{booktabs}
\usepackage{makecell}
\usepackage{amssymb}
\usepackage{bm}		
\usepackage[table]{xcolor}
\usepackage[colorlinks, citecolor=blue]{hyperref} 
\usepackage{threeparttable}
\usepackage{subfigure}
\usepackage{caption}

\usepackage{colortbl}
\usepackage{nicematrix}
\usepackage{tabularx}
\usepackage{multirow}
\usepackage{dsfont}

\usepackage[american]{babel}

\def\onedot{.}
\def\eg{\emph{e.g}\onedot}
\def\ie{\emph{i.e}\onedot}

\newcolumntype{L}[1]{>{\raggedright\arraybackslash}p{#1}}
\newcolumntype{C}[1]{>{\centering\arraybackslash}p{#1}}
\newcolumntype{R}[1]{>{\raggedleft\arraybackslash}p{#1}}

\usepackage{amsthm}

\begin{document}
%
\title{
Learning Probabilistic Coordinate Fields \\
for Robust Correspondences}

\author{Weiyue~Zhao, 
	Hao~Lu,~\IEEEmembership{Member,~IEEE},
	Xinyi~Ye,
	Zhiguo~Cao,~\IEEEmembership{Member,~IEEE},
	and~Xin~Li,~\IEEEmembership{Fellow,~IEEE}
\thanks{This work is supported by the National Natural Science Foundation of China under Grant No. 62106080. Corresponding author: Z. Cao.}

\thanks{W. Zhao, H. Lu, X. Ye and Z. Cao are with Key Laboratory of Image Processing and Intelligent Control, Ministry of Education; School of Artificial Intelligence and Automation, Huazhong University of Science and Technology, Wuhan, 430074, China (e-mail: \{zhaoweiyue,hlu,xinyiye,zgcao\}@hust.edu.cn).}
\thanks{X. Li is with the Lane Department of Computer Science and Electrical Engineering, West Virginia University, Morgantown WV 26506-6109 (e-mail: xin.li@ieee.org).}
}

\markboth{Manuscript Submitted to IEEE Trans. Pattern Analysis \& Machine Intelligence; Jun~2022}%
{Shell \MakeLowercase{\textit{zhao et al.}}: Learning Probabilistic Coordinate Fields for
Robust Correspondences}
%



\IEEEtitleabstractindextext{%
\begin{abstract}
We introduce Probabilistic Coordinate Fields (PCFs), a novel geometric-invariant coordinate representation for image correspondence problems. In contrast to standard Cartesian coordinates, PCFs encode coordinates in correspondence-specific barycentric coordinate systems (BCS) with affine invariance. To know \textit{when and where to trust} the encoded coordinates, we implement PCFs in a probabilistic network termed PCF-Net, which parameterizes the distribution of coordinate fields as Gaussian mixture models. By jointly optimizing coordinate fields and their confidence conditioned on dense flows, PCF-Net can work with various feature descriptors when quantifying the reliability of PCFs by confidence maps. An interesting observation of this work is that the learned confidence map converges to geometrically coherent and semantically consistent regions, which facilitates robust coordinate representation. By delivering the confident coordinates to keypoint/feature descriptors, we show that PCF-Net can be used as a plug-in to existing correspondence-dependent approaches. Extensive experiments on both indoor and outdoor datasets suggest that accurate geometric invariant coordinates help to achieve the state of the art in several correspondence problems, such as sparse feature matching, dense image registration, camera pose estimation, and consistency filtering. Further, the interpretable confidence map predicted by PCF-Net can also be leveraged to other novel applications from texture transfer to multi-homography classification. 
\end{abstract}

}

\maketitle

\IEEEdisplaynontitleabstractindextext

%
\IEEEpeerreviewmaketitle

\IEEEraisesectionheading{\section{Introduction}\label{intro}}

\begin{figure}[!t]
  \centering
  \includegraphics[width=\linewidth]{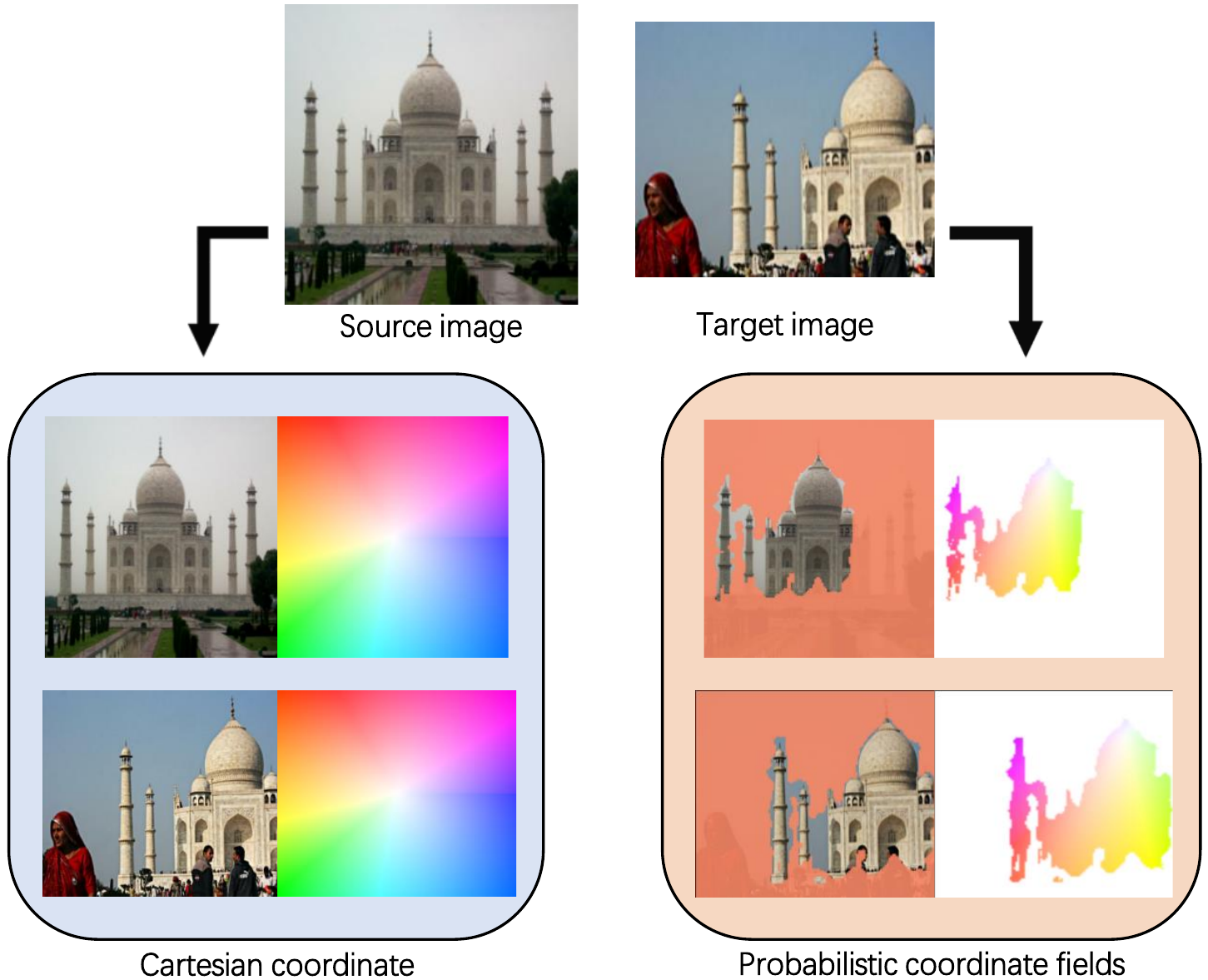}
    \caption{\textbf{Comparison of coordinate encoding approaches between an image pair.} Coordinates are color-coded. Between image pairs, the same color indicates the same coordinate representation. Our probabilistic coordinate fields (PCFs) exhibit two distinct advantages over conventional coordinate systems such as the Cartesian coordinate system: \romannumeral1) PCFs are geometrically invariant. The Cartesian coordinates remain unchanged even if the temple is horizontally displaced from source to target, while only geometrically consistent positions take the same coordinate representation in PCFs. \romannumeral2) PCFs aim to encode reliable regions. Our approach searches for confident maps (orange masks) used to reject occluded regions (\eg, walking pedestrians) and irrelevant regions (\eg, the sky) to preserve semantic consistency.}
\label{pic:fig1}
\end{figure}

\IEEEPARstart{T}{he} search for robust correspondences is a fundamental problem in computer vision and can benefit a number of applications, such as image registration~\cite{truong2019glampoints,zhang2019learning}, 3D reconstruction~\cite{schonberger2016structure,wu2011visualsfm}, and image fusion~\cite{jiang2020multi}. Many image matching approaches~\cite{sun2020acne,yi2018learning,zhao2019nm} resort to local descriptors to represent features around the keypoints of neighboring regions of interest~\cite{arandjelovic2012three,detone2017toward,detone2018superpoint,lowe2004distinctive,rublee2011orb}. Due to their local nature, 
few works~\cite{sarlin2020superglue,gehring2017convolutional} embed keypoint coordinates into descriptors to improve feature matching. Although some approaches 
can overcome changes in illumination and blur distortion, they cannot tackle challenging scenarios such as repeated patterns and low-texture regions. In fact, standard coordinates, such as the Cartesian coordinate shown in Fig.~\ref{pic:fig1}(a), are sensitive to affine transforms. They offer poor positional information or geometric consistency between image pairs. Recently, several works established dense matches in an end-to-end manner~\cite{jiang2021cotr,rocco2020efficient,sun2021loftr,tyszkiewicz2020disk}. However, they 
suffer from the same problem as those of improper coordinate encoding, such as image-level multi-frequency sine/cosine embedding. Most recently, it was pointed out in LoFTR~\cite{sun2021loftr} that inappropriate positional encoding can even significantly degrade performance.

Unlike previous approaches, we propose to encode image coordinates based on \emph{barycentric coordinate system} (BCS)~\cite{marschner2018fundamentals}. This system has been widely used in computer graphics due to its geometric invariance under rigid transforms.  
By constructing a pair of correspondence-specific coordinate systems between two images, we can acquire barycentric coordinate fields (BCFs) to facilitate some correspondence problems. BCFs are not only affine invariant but also geometrically coherent between image pairs. However, the geometric coherence of BCFs is prone to errors in the presence of occlusion, perspective/non-rigid transforms, or inaccurately constructed BCSs. Therefore, it is crucial to know \emph{when and where} to trust the encoded coordinate fields \cite{truong2021learning}, regardless of whether they are Cartesian or barycentric. 

Inspired by the estimation of uncertainty in optical flow and dense geometry matching~\cite{gast2018lightweight,kendall2017uncertainties,truong2021learning}, we propose to predict the confidence values for BCFs under the supervision of estimated flows. Specifically, we first introduce the notion of Probabilistic Coordinate Field (PCF), an uncertainty-aware geometric-invariant coordinate representation. To generate the PCF, we then develop a network termed PCF-Net that jointly optimizes coordinate encoding and confidence estimation. The key idea behind PCF-Net is to generate correspondence-specific BCSs from dense flows and, more importantly, to use the estimated flows to supervise the encoding of the BCFs such that geometric coherence and semantic consistency can be preserved. To find reliable co-visible regions between paired BCFs, we parameterize the distribution of BCFs as Gaussian mixture models and infer the parameters of probabilistic models using PCF-Net when forming the confidence maps. In this way, PCFs can be interpreted as {\em confidence maps} over BCFs, as shown in Fig.~\ref{pic:fig1}.

We have evaluated our approach on indoor (SUN3D~\cite{xiao2013sun3d} and ScanNet~\cite{dai2017scannet}) and outdoor (YFCC100M~\cite{thomee2016yfcc100m}, PhotoTourism~\cite{jin2020image}, and MegaDepth~\cite{2018MegaDepth}) datasets on top of various state-of-the-art matching approaches (SuperGlue~\cite{sarlin2020superglue}, LoFTR~\cite{sun2021loftr}, and OANet~\cite{zhang2019learning}). We first pre-train the PCF-Net using a mixture of synthetic and realistic datasets, and then incorporate the fixed PCF-Net into different baselines to replace the original coordinate representation with PCFs. For detector-based approaches (SuperGlue and OANet), we also tested them under different descriptors, including RootSIFT~\cite{arandjelovic2012three}, HardNet~\cite{mishchuk2017working}, and SuperPoint~\cite{detone2018superpoint}. Extensive experiments show that PCFs are generic coordinate representations for correspondence problems. PCFs can be used as a plug-in to help advance the state-of-the-art and, therefore, to benefit a number of correspondence-dependent tasks, such as sparse feature matching, dense image registration, camera pose estimation, and consistency filtering. Meanwhile, we have also discovered some exclusive properties of the confidence maps predicted by PCF-Net; for instance, they could be useful for other tasks ranging from texture transfer to multi-homography classification.

\vspace{0.1in}
\noindent\textbf{Summary of Contributions.}
We show that geometric-invariant coordinate representations can be powerful for image correspondence problems. Technically, we introduce PCF-Net to encode geometric-invariant coordinates with confidence estimation. To our knowledge, our work is the first attempt to unify \emph{robust} feature correspondences with \emph{probabilistic} coordinate encoding. As shown by our experimental results, the proposed PCF is versatile, supporting a variety of keypoint descriptors and matching algorithms. The PCF-Net can be applied as a plug-in module in existing baselines with affordable computational overhead (on average $116$ ms on a modern GPU). Furthermore, the interpretable confidence maps predicted by PCF-Net for coordinate representations can be exploited for other novel applications. 

\begin{figure*}[h]
\centering
    \includegraphics[width=1.0\textwidth]{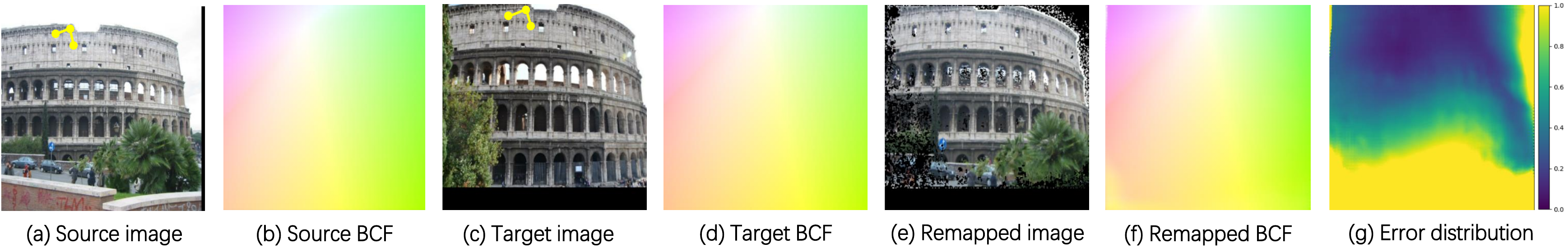}
\caption{\textbf{Visualization of correspondence-specific BCS and the error distribution of the target coordinate field.} Given images (a) and (c), we encode the corresponding barycentric coordinate fields (BCFs) (b) and (d), based on a pair of correspondence-specific coordinate systems (BCSs, yellow arrows in (a) and (c)). (a) and (b) are remapped to (e) and (f) with ground-truth flows. (g) compares the difference between (d) and (f) (nonmatched regions marked yellow).}
\label{fig:vis}
\end{figure*}

\section{Related Work}

\vspace{5pt}
\noindent\textbf{Feature Matching.}
In general, feature matching can be categorized into detector-based and detector-free matching. Detector-based approaches adopt mainly hand-crafted local descriptors~\cite{arandjelovic2012three,lowe2004distinctive,rublee2011orb} or learning-based  descriptors~\cite{detone2017toward,detone2018superpoint,mishchuk2017working}. 
The local-only descriptors~\cite{mishchuk2017working,rublee2011orb} encode local patterns and are typically extracted from the image patches. However, when repeated patterns appear, they may not work well. The local-global descriptors~\cite{arandjelovic2012three,detone2018superpoint,lowe2004distinctive} simultaneously encode local and global cues. Global cues can be discriminative even with similar local patterns. However, regardless of the descriptor type, the inclusion of global features is limited to gray or color information, ignoring complex textures. 
Furthermore, despite detector-free approaches~\cite{jiang2021cotr,li2020dual,Rocco18b,sun2021loftr} that work directly on dense feature maps, the limited resolution of feature maps can still cause information loss. In this work, we use geometrically invariant coordinates to characterize spatial locations between paired images. The generated coordinate fields provide geometrically and semantically consistent information regardless of the type of descriptor and the resolution of the image.

\vspace{5pt}
\noindent\textbf{Positional Embedding.}
Positional representation has recently received a lot of interest. The absolute position encoding with the multifrequency sine/cosine function was first introduced by~\cite{vaswani2017attention}. 
\cite{carion2020end,sun2021loftr} also implement similar encoding strategies to address object detection and local feature matching. By incorporating position encoding into local descriptors, features are expected to be position-dependent. Another choice is the learned positional encoding~\cite{gehring2017convolutional,sarlin2020superglue}. In addition to the absolute position, recent work also considers relative positional encoding~\cite{DBLP,wu2021rethinking,edstedt2022deep}. However, all of these positional representations are sensitive to geometric transforms. Especially for image matching, there is no explicit geometric coherence between pairwise positional representations. Instead, we propose to encode geometric invariant coordinates based on BCSs, which have been shown to greatly benefit geometric coherence between paired images.

\vspace{5pt}
\noindent\textbf{Uncertainty Estimation.}
Compared to other techniques, uncertainty estimation is less studied in feature matching. DGC-Net~\cite{melekhov2019dgc} proposes a matchability decoder to remove uncertain correspondences. \cite{shen2020ransac} attempts to improve the accuracy of correspondence matching by adding probability maps. Similarly, some optical flow~\cite{gast2018lightweight,ilg2018uncertainty} and geometric dense matching~\cite{gal2016dropout,kendall2017uncertainties,truong2021learning} approaches also predict the confidence map when estimating a flow map. In this work, we need to address how to find confident regions in the PCFs because barycentric coordinate systems may not be accurate. 
However, compared with the smooth probabilistic map generated by flow models, we prefer instead a more distinctive probabilistic map with clear demarcation to classify reliable/unreliable coordinate regions.

\section{Background, Motivation, and Overview} \label{sec:background and motivation}
 


\subsection{Background: Barycentric Coordinates Fields}
For explicit representation of position, coordinates are widely used within convolutional networks~\cite{jin2020image,sun2020acne,zhao2019nm} or with multifrequency sine/cosine functions~\cite{sun2021loftr,vaswani2017attention}. However, most coordinate representations suffer from geometric transforms. In the context of correspondence problems, a key insight of our work is that one should use geometric-invariant coordinates, such as BCS. 
Within a BCS~\cite{marschner2018fundamentals}, barycentric coordinates $(\lambda_1, \lambda_2, \lambda_3)$ of a point $P$ are defined by
\begin{equation} \label{eq:bary_coor} \small
    \setlength\abovedisplayskip{5pt} 
    \setlength\belowdisplayskip{5pt} 
    \lambda_1=\frac{S_{\triangle PBA}}{S_{\triangle ABC}}\,,
    \lambda_2=\frac{S_{\triangle PAC}}{S_{\triangle ABC}}\,,
    \lambda_3=\frac{S_{\triangle PCB}}{S_{\triangle ABC}}\,,
\end{equation}
where $\lambda_1 + \lambda_2 + \lambda_3 = 1$, and $S_{\triangle}$ represents the triangle area (the detailed review of BCS can be found in Appendix~A).


Given a BCS, we can form the image coordinates to be a barycentric coordinate field (BCF). According to Eq.~\eqref{eq:bary_coor}, there is a linear relation between $\lambda_1$, $\lambda_2$, and $\lambda_3$ such that two out of three can represent the rest. Without loss of generality, we choose $(\lambda_1, \lambda_2)$ as the coordinate representation. 
Formally, given an image pair $Z = (I^s, I^t)$ where $I^s, I^t\in \mathbb{R}^{H \times W \times 3}$, we can construct a pair of BCFs $(C^s, C^t)$ from the image coordinates using two sets of BCS, respectively, where $C^s$ is the source BCF and $C^t$ is the target. This process defines a transformation $\mathcal{F}:\mathbb{R}^{H \times W \times 2} \rightarrow \mathbb{R}^{H \times W \times 2}$ by Eq.~\eqref{eq:bary_coor}, where $H$ and $W$ are the height and width of the image, respectively.
Note that any single coordinate system alone does not provide information on the geometric coherence between pairs of images.

\subsection{Motivation: Confidence Over Coordinate Fields} \label{subsec:Barycentric Coordinate Field}
A key step of our work is to bridge $C^s$ and $C^t$ using correspondences, as shown in Figs.~\ref{fig:vis}(a)-(d). In particular, each vertex of the target BCS is associated with a correspondence and has a matched source vertex. Ideally, if geometric transforms between image pairs are constrained to affine transforms, the geometric coherence of the paired BCFs is highly reliable due to the invariance property of the BCS (please refer to the Appendix~A for more details about the affine invariant property of BCS). However, occlusions, non-rigid transforms, or large displacements often appear in reality. In these cases, the matched geometric coherence is prone to errors. This can be observed from the coordinate error map (see Fig.~\ref{fig:vis}(g)) between $C^t$ (see Fig.~\ref{fig:vis}(d)) and $C^r$ (see Fig.~\ref{fig:vis}(f)), where $C^r$ is the remapped coordinate field.

From the error map, we can make the following important observation: errors are subject to a consistent distribution locally, but show a trend of smooth transition globally. In general, errors increase approximately linearly with increasing distance from the origin of the BCS (the white point in Fig.~\ref{fig:vis}(f)). Since BCFs could be erroneous, it is critical to know when and where to trust the encoded coordinate field. Thus, we present probabilistic coordinate fields (PCFs), where a confidence metric is introduced into the BCFs and modeled in a probabilistic framework. Next, we formally define correspondence-specific BCSs, introduce PCFs by conditional modeling, and show how PCFs can be generated from a network.

\subsection{Overview of Our Method}

As illustrated in motivation and Fig.~\ref{pic:fig1}, our objective is to generate a set of geometry-invariant coordinate fields with confidence maps for a given image pair. To achieve this goal, we have developed a new model called PCF-Net which utilizes affine-invariant barycentric coordinates. This architecture takes the RGB image pair as input and produces the barycentric coordinate fields and the corresponding pixel-level confidence values. By combining the coordinate representations with their respective confidence values, we can generate the probabilistic coordinate field.

Specifically, we introduce the barycentric coordinate system and analyze its limitations in this Section. In Section~\ref{sec: From Barycentric to Probabilistic Coordinate Fields}, we explain how to build barycentric coordinate systems between an image pair and mathematically model uncertainty estimation for coordinate systems. In Section~\ref{sec:pcfnet}, we provide technical details of PCF-Net, which implements uncertainty estimation with Gaussian mixture models. In Section~\ref{sec:strategies}, we explain in detail how to incorporate our approach into various correspondence problems.


\begin{figure*}[!t]
\centering  
\includegraphics[width=0.95\textwidth]{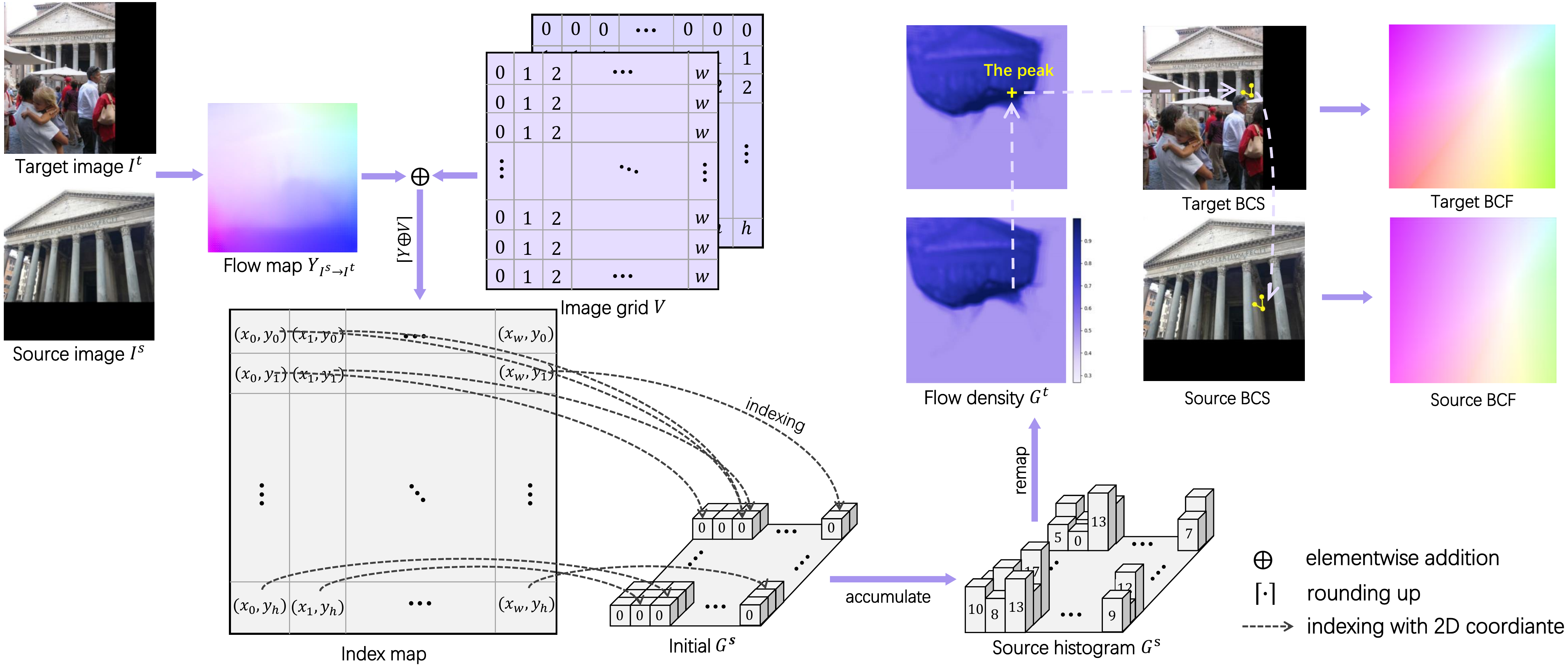}
\caption{\textbf{Pipeline of generating correspondence-specific BCSs.} The index map is obtained by summing the flow map $Y_{I^s \rightarrow I^t}$ and an image coordinate grid $V$. We accumulate the mapping times of the pixels of the source image indexed by the index map and obtain the source histogram $G^s$. Then $G^s$ is remapped to the target flow density $G^t$ according to $Y$. The peak of $G^t$ (``the peak'') is selected as the origin of the target BCS. Then we can construct the correspondence-specific source BCS using the correspondences given by $Y$. The BCSs are colored yellow in the source and target images. On the basis of BCSs, we obtain the corresponding BCFs.}
\label{fig:density_map}
\end{figure*}


\section{From Barycentric to Probabilistic Coordinate Fields} \label{sec: From Barycentric to Probabilistic Coordinate Fields}

Here we present our general idea of how to build the BCSs from an image pair, how to generate BCFs conditioned on BCSs, and how to model uncertainty into BCFs.

\subsection{Barycentric Coordinate Systems From Flows} \label{sec:select_origin}
Given a pair of images, the correspondence problem refers to the problem of determining which parts of one image correspond to which parts of the other image. 
Correspondences can be generated from sparse/dense feature matching. However, sparse matching is often limited by local features or sparse keypoints, whereas dense matching usually requires expensive computation and inference. Instead, optical flows \cite{beauchemin1995computation} often provide a dense but coarse correspondence map at low resolution (LR) at low computational cost. This conforms to the construction of initial BCFs when precise correspondences are not necessary. 

In this work, we follow the recent GLU-Net~\cite{truong2020glu}, a unified model capable of geometric matching and optical flow estimation, to acquire an initial set of correspondences. In particular, given an image pair $Z$, GLU-Net generates a low-resolution (LR) flow map $Y_{I^s \rightarrow I^t} \in \mathbb{R}^{H_L \times W_L\times 2}$, where $H_L = \frac{H}{4}$ and $W_L = \frac{W}{4}$. To simplify the notation, we use $Y$ in what follows. In particular, we choose the set of correspondences $S$ from the flow map $Y$, where $S$ is made up of three paired correspondences. Using $S$, we can construct a pair of correspondence-specific BCSs between $Z$. According to Eq.~\eqref{eq:bary_coor}, the initial BCFs $(C^s$, $C^t)$ can be derived accordingly.

\vspace{5pt}
\noindent\textbf{Correspondence-Specific BCSs}. 
Here, we explain how to set up the correspondence between specific BCSs (as shown in Fig.~\ref{fig:density_map}).
Given the flow map $Y$ from the source image to the target and an image coordinate grid $V$, a source histogram $G^s$ can be obtained by $\forall p \in \lceil Y \oplus V \rceil, G^s(p) = G^s(p) + 1$, where $\oplus$ denotes elementwise addition and $\lceil \cdot \rceil$ denotes rounding operation. Specifically, $\lceil Y \oplus V \rceil$ obtains an index map, each value of which represents the coordinates of the corresponding point in the source image that corresponds to the current index point. Then, we initialize an all-zero histogram $G^s$ and accumulate it indexed by the index map. The value of the source histogram $G^s$ indicates the number of times a pixel of the source image is mapped to the target image. Then the flow map $Y$ can remap $G^s$ to the target $G^t$, denoted by the flow density map, which characterizes the mapping frequency of the pixels of the source image in the flow distribution.
After applying the average pooling on $G^t$ with a kernel size of $K$ ($K$ is a positive odd integer), the peak (yellow marked in Fig.~\ref{fig:density_map}) of the average density map is chosen as the origin of the target BCS. The other two vertices are randomly chosen around the origin, with a radius of $(K-1)/2$. Once the target BCS is built, the corresponding source BCS can be constructed according to the flow map and the target BCS. 

Since $Y$ is of low resolution, we upsample $C^s$ and $C^t$ with bilinear interpolation to apply them to images of the original resolution. It is worth noting that upsampling does not affect the geometric coherence of coordinate fields due to the scale invariance of barycentric coordinates. Nevertheless, the BCFs are still not ready for use because the flow map can be inaccurate, such that the generated correspondence set and constructed BCSs are inaccurate. Hence, it is important to know which correspondences are trustable on the flow map.

\subsection{Probabilistic Coordinates via Conditional Modeling} 
\label{subsec:Coordinates in Conditional Modeling}
To find reliable geometrically coherent regions, our idea is to design a network to predict a confidence map for BCFs conditioned on the flow map. Intuitively, this idea can be implemented by designing the network to predict the conditional probability used to represent the confidence of the coordinate fields. 
Given an image pair $Z$, coupled BCFs $(C^s, C^t)$, and flow map $Y$, our goal is to generate a confidence map $M \in \mathbb{R}^{H \times W}$ relating $C^s$ to $C^t$. This can be achieved by defining a conditional probability density $\mathcal{P}(X(C^s;Y)|\Psi(\theta))$, where $X(C^s;Y)$ denotes the remapped coordinate field $C^s$ conditioned on the flow $Y$, and $\Psi(\theta)$ is a network parameterized by $\theta$ that predicts the parameters of the probabilistic model. If assuming spatial independence, for any position $(i, j)$, $\mathcal{P}(X|\Psi)$ amounts to a family of distributions $\prod_{ij}\mathcal{P}(x_{ij}|\psi_{ij})$, where $x_{ij}$ denotes the remapped barycentric coordinate depending on the optical flow. $\psi_{ij}$ is a network that outputs the parameters of the probability density specific to the spatial location $(i, j)$. To simplify the notation, we eliminate the subscripts $i,j$ in the following expressions.

A common choice to model a conditional probability density is to use a Laplacian/Gaussian distribution~\cite{gast2018lightweight,ilg2018uncertainty,kendall2017uncertainties,walz2020uncertainty,truong2021learning}. Their main difference lies in the definition of loss function: $\ell_1$ loss $|x-\mu|$ vs. $\ell_2$ loss $(x-\mu)^{2}$, where $x$ is a variable and $\mu$ is the mean. Interestingly, many optical flow approaches adopt the Laplacian model due to the robustness of $\ell_1$~\cite{ilg2018uncertainty,truong2021learning} (i.e., $\ell_1$ distance does not significantly penalize large flow errors). However, in our correspondence problems, we expect an opposite behavior-~\ie, large coordinate errors should be penalized more because they directly reflect the accuracy of the correspondence. In other words, unlike probabilistic flow models~\cite{ilg2018uncertainty,truong2021learning}, our model must be sensitive to large coordinate errors to acquire a discriminative confidence map. An experimental justification for the preference for $\ell_2$ over $\ell_1$ can be found in our ablation studies (Section~\ref{subsec:experi_probablistic}).

\section{Probabilistic Coordinate Fields: BCFs on Confidence} \label{sec:pcfnet}

\begin{figure*}[!t] 
    \centering
    \includegraphics[width=0.98\textwidth]{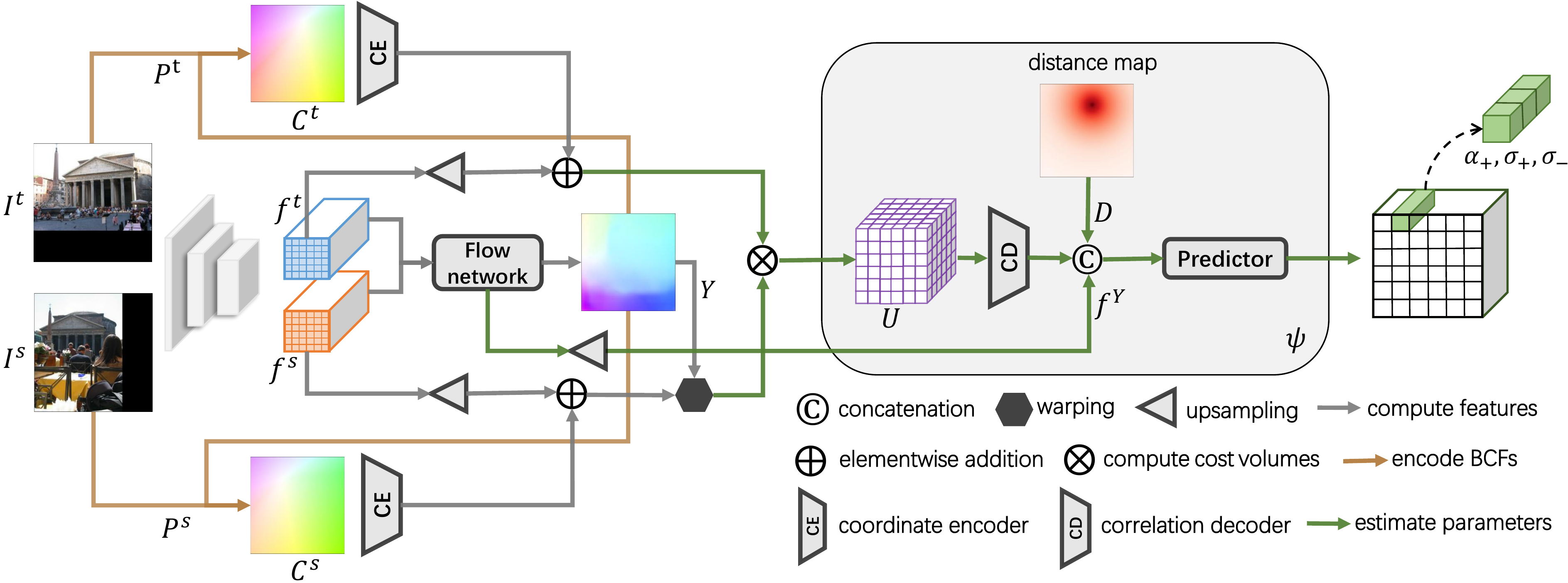}
    \caption{\textbf{Technical pipeline of PCF-Net.} We first extract features $f^s$ and $f^t$ using a pretrained CNN pyramid from an image pair. Given the flow map $Y$ from the flow network, we then build a couple of BCFs $B=(C^s, C^t)$ from the image coordinates ($P^s, P^t$). The BCFs are integrated with the upsampled features through a shared coordinate encoder (CE) to compute a correlation map $U$ following~\cite{truong2021learning}, also with the flow map $Y$. Finally, we concatenate the flow feature $f^Y$, a distance map $D$, and the output of the correlation decoder (CD) to predict the parameters of the probabilistic models. Note that our model $\psi$ outputs the parameters ${\alpha_+, \sigma_+, \sigma_-}$ of probabilistic distribution for each pixel location. The details of architecture can be found in Appendix~B.}
\label{fig:architecture}
\end{figure*}

The goal of probabilistic modeling of coordinate fields is to use a probabilistic model to identify reliable regions in BCFs. In practice, we first need the model to generate a confidence map, which is the core of Probabilistic Coordinate Fields (PCFs). Then we show how to generate the PCF with a network termed PCF-Net, which extends the previous work PDC-Net~\cite{truong2021learning}.

\subsection{Probabilistic Modeling by Gaussian Mixture Models} \label{subsec:Coordinates as Gaussian Mixture Models}

We assume that every spatial location obeys a 2D Gaussian distribution $\mathcal{G}(x)\sim N(\mu_u, \sigma_u^2)\cdot N(\mu_v, \sigma_v^2)$, where each target coordinate $x=(u, v) \in \mathbb{R}^2$ is modeled by two conditionally independent 1D Gaussians: $N$($\mu_u$, $\sigma_u^2$) and $N$($\mu_v$, $\sigma_v^2$) respectively.
Additionally, we assume equal variances on both coordinate axes such that $\sigma_u^2 = \sigma_v^2 = \sigma^2$. By further defining $\mu = [\mu_u, \mu_v]^T$, the 2D Gaussian distribution $\mathcal{G}$ amounts to
\begin{equation} \label{eq:guassian-simple} \small
    \mathcal{G}(x|\mu, \sigma^2) = \frac{1}{2\pi\sigma^2}e^{-{\frac{{||x-\mu||}_2^2}{2\sigma^2}}} \,.
\end{equation}

In theory, we expect that the above probabilistic regression model can fit the empirical distribution shown in Fig.~\ref{fig:vis}(g) as well as possible. Inspired by ~\cite{kendall2017uncertainties}, we choose Gaussian Mixture Models (GMMs) for flexible modeling. Concretely, we consider a model consisting of $k$ Gaussian components, \ie, $\mathcal{P}(x|\psi) = \sum_{n=1}^{k} \alpha_n\mathcal{G}(x|\mu, \sigma_n^2)$, where all $K$ components share the same $\mu$ but different variances $\sigma_n$, and $\mu$ is given by the remapped source coordinate field $C^r$ (refer to Fig.~\ref{fig:vis}(f)). 
$\alpha_n \geq 0$ satisfies $\sum_{n=1}^{k}\alpha_n = 1$. Empirically, when $K$ increases, a GMM should fit complex distributions better. However, we experimentally find that a complex model can cause overfitting in some uncertain regions, causing small $\sigma_n^2$ with large $\alpha_n$. Recall that our goal is to discriminate the coordinate field into reliable/unreliable regions. Inspired by PDC-Net~\cite{truong2021learning}, we propose to use a constrained two-component GMM to model the distribution, defined by
\begin{equation} \label{eq:GCM}\small
    \mathcal{P}(x|\psi) = \alpha_+\mathcal{G}(x|\mu, \sigma_+^2) + (1 - \alpha_+)\mathcal{G}(x|\mu, \sigma_-^2) \,,
\end{equation}
where $(\alpha_+, \sigma_+^2, \sigma_-^2)$ are all predicted by the network $\psi(\theta)$. For robust fitting, we add some constraints to both variances $(\sigma_+^2$ and $\sigma_-^2)$ and $\alpha_+$.
Unlike~\cite{truong2021learning}, we applied different constraint strategies to meet the modeling needs of the coordinate probability model. The details are discussed in the following. An experimental comparison of our preference over PDC-Net can be found in Section~\ref{subsec:experi_probablistic}.

\vspace{5pt}
\noindent\textbf{Constraint on $\sigma_+^2$ and $\sigma_-^2$}.
Intuitively, each variance $\sigma_n$ in standard GMMs accounts for a certain range of uncertainties, loosely corresponding to different error regions in Fig.~\ref{fig:vis}(g). To identify reliable regions more explicitly, we constrain $\sigma_+^2$ and $\sigma_-^2$ into different ranges such that
\begin{equation} \label{eq:constrained sigma}\small
    0 \leq \sigma_+^2 \leq \delta_+ < \delta_++\Delta\delta \leq \sigma_-^2 < \delta_- \,,
\end{equation}
where $\sigma_+^2$ accounts for reliable regions and $\sigma_-^2$ for erroneous regions. $\delta_+$ and $\delta_-$ are empirically set hyperparameters in this work. The margin $\Delta\delta$ is used to avoid a smooth transition and to force the network to make a choice at every spatial location.
We also find that it is useful to constrain the value of $\sigma_+^2$ to $\delta_+$, rather than using a fixed $\sigma_+^2$ as in~\cite{truong2021learning}. 

\vspace{5pt}
\noindent\textbf{Constraint on $\alpha_+$}.
In standard GMMs, each $\alpha_n$ controls the contribution of a component. In the open literature, some approaches~\cite{danelljan2020probabilistic,gast2018lightweight,ilg2018uncertainty} predict unconstrained $\alpha_n$'s independently and then use a ${\tt\small softmax}$ layer for normalization. However, a potential issue is that there is no information interaction or explicit value constraints between different $\alpha_n$'s. This may cause confusion, so similar $\alpha_n$'s are used in different components, which is undesirable for discriminating reliable/unreliable regions. Instead of predicting $\alpha_n$ for each component, here we only predict $\alpha_+$ and set $\alpha_-=1-\alpha_+$. In this way, the network is required to make a hard choice between two components. Note that we also normalize $\alpha_+$ with a function ${\tt\small sigmoid}$.

\begin{figure}[t]
\centering  
\includegraphics[width=0.49\textwidth]{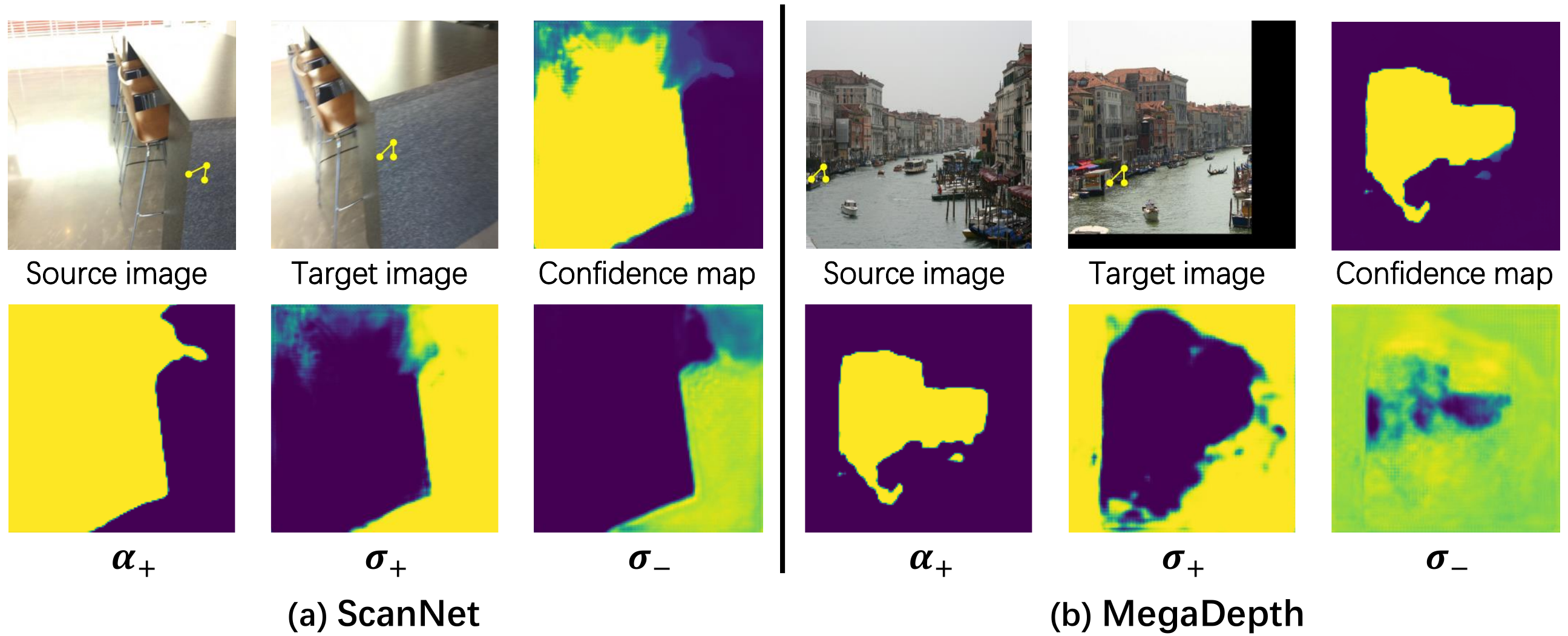}
\caption{\textbf{Detailed visualization of various parameters on multiple pairs of images.} The barycentric coordinate systems (BCSs) are marked by {{yellow}} ($\wedge$-shape) in the source and target images. The parameters $\alpha_+$, $\sigma_+$, and $\sigma_-$ are predicted by our PCF-Net network. The confidence map of the target image is calculated using the probabilistic model.} 
\label{fig:two dataset}
\end{figure}


\begin{figure*}[!t]
\centering  
\includegraphics[width=1.0\textwidth]{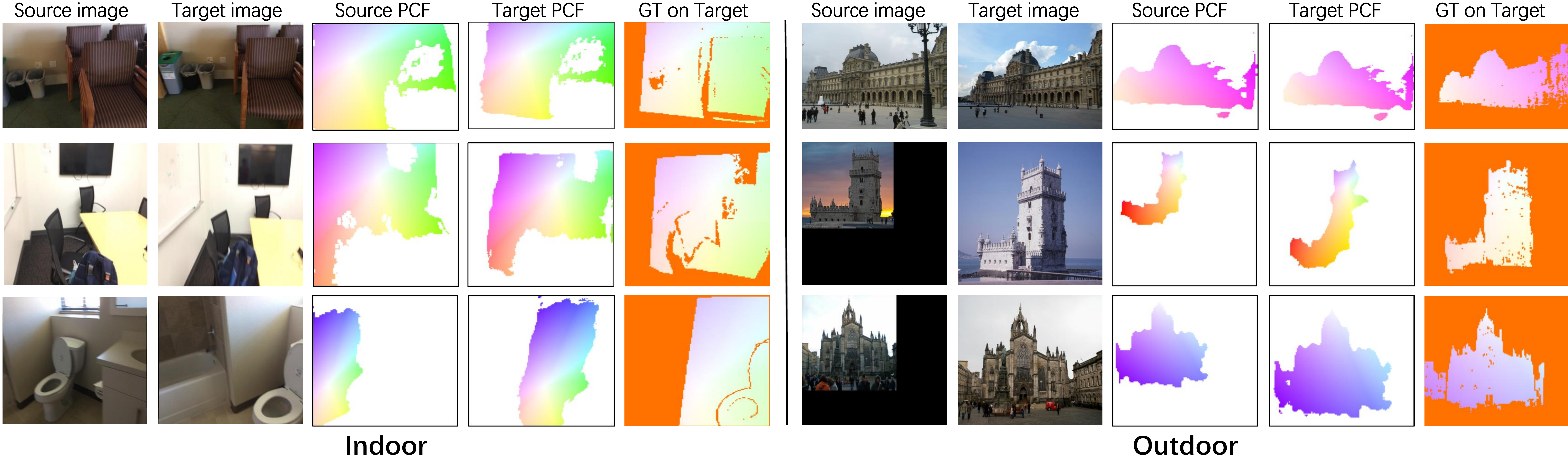}
\caption{\textbf{Qualitative examples of our approach PCF-Net on both MegaDepth and ScanNet datasets.} In the $3^{rd}$ and $4^{th}$ columns, we visualize the probabilistic coordinate fields (PCFs) of the source and target, respectively. White indicates unreliable regions according to the confidence map. In the last column, we warp the ground-truth flow into the target image. Orange indicates that no matching correspondence between image pairs can be found.} 
\label{fig:vis_pcf}
\end{figure*}


\vspace{5pt}
\noindent\textbf{Confidence Map $M$}.
Given the predicted parameters $(\alpha_+, \sigma_+^2, \sigma_-^2)$ of the probabilistic model in Eq.~\eqref{eq:GCM}, we can compute the confidence $M_{ij}$ at location $(i,j)$ within a radius ${||x-\mu||}_2<R$ as
\begin{equation}\small \label{eq:confidence map}
    \begin{split}
        M_{ij} &= \int_{\{x\in\mathbb{R}^2:{||x-\mu||}_2<R\}}\mathcal{P}(x|\psi) dx \\
        &=\int_{-R}^{R}\int_{-R}^{R}\alpha_+\mathcal{G}(x|\mu, \sigma_+^2) + (1 - \alpha_+)\mathcal{G}(x|\mu, \sigma_-^2)dudv \\
        & =1 - \exp({-\frac{R^2}{2\sigma_-^2}}) +\alpha_+\left[\exp({-\frac{R^2}{2\sigma_-^2}}) - \exp({-\frac{R^2}{2\sigma_{+}^2}})\right] \,.
    \end{split}
\end{equation}
Finally, once $M$ is obtained, it can be used to interact with the remapped coordinate field $C^r$ to generate the PCF, given by $M \otimes C^r$, where $\otimes$ is the element-wise multiplication. Note that the PCF has a hard form and a soft form, conditioned on whether the map $M$ is binarized.


\subsection{Probabilistic Coordinate Field Network}
\label{subsec:PCF-Net}
Here, we show how to generate the PCF with a network called the Probabilistic Coordinate Field Network (PCF-Net). The PCF-Net pipeline is shown in Fig.~\ref{fig:architecture}. Mostly, it includes a coherence module and a probabilistic model predictor.

\vspace{5pt}
\noindent\textbf{Coherence Module}. 
The coherence module aims to construct geometric and semantic coherence between an image pair ($I^s, I^t$) before predicting the model parameters. We model geometric and semantic coherence by combining the coordinate features $(C^s, C^t)$  with the image features $(f^s, f^t)$. The coordinate features extracted by a shared coordinate encoder are integrated into the upsampled content features. By warping the integrated source features with the flow map $Y$, a 4D correlation map $U$~\cite{truong2021learning} can be calculated for parameter estimation.

\vspace{5pt}
\noindent\textbf{Probabilistic Model Predictor}.
To predict the parameters of the probabilistic model, we also integrate the information of a prior distribution and the flow map. As shown in Fig.~\ref{fig:vis}(g), we propose to use a distance map $D$ to approximate the error distribution, given by
\begin{equation} \label{eq:distance map}\small
    \setlength\abovedisplayskip{5pt} 
    \setlength\belowdisplayskip{5pt} 
    D_{ij} = \exp(-\gamma \cdot\frac{1}{d_{ij}})\,,
\end{equation}
where $\gamma$ controls the decrease speed and $d_{ij}$ is the distance between a location $(i,j)$ and the origin of BCS. In addition, PDC-Net~\cite{truong2021learning} points out that dense flow estimation is important to move objects independently. Similarly to $D$, the flow characteristic $f^Y$ is also considered. Finally, we concatenate them and feed them to a predictor to regress the parameters of the probabilistic model through several convolutional layers (reference to the implementation in~\cite{truong2021learning}).


\vspace{5pt}
\noindent\textbf{Loss Function}. Using the following practices in probabilistic regression tasks~\cite{ilg2018uncertainty,kendall2017uncertainties,varamesh2020mixture}, we train our model with the negative log-likelihood loss function. 
For a given image pair, the coupled coordinate fields $(C^s, C^t)$ and the ground truth flow $Y_{gt}$, the loss takes the form of
\begin{equation}\small
    \setlength\abovedisplayskip{5pt} 
    \setlength\belowdisplayskip{5pt} 
    \mathcal{L} = -\log{\mathcal{P}(\hat{X}(C^s;Y_{gt})|\Psi(\theta))} \,,
\end{equation}
where $\hat{X}(C^s;Y)$ denotes the remapped coordinate field $C^s$ conditioned on the flow $Y_{gt}$.
Note that this loss function explicitly supervises the reliable regions of PCFs. Similarly, the remapped coordinate field $\hat{X}$ also implicitly supervises the learning of the flow map.

\subsection{Visualization of PCFs}
To better understand our probabilistic model, we visualize the parameters predicted by PCF-Net, as shown in Fig.~\ref{fig:two dataset}. Specifically, we choose indoor and outdoor scenes to demonstrate the generalization of our approach. The inferred parameters are all predicted by PCF-Net for the target image. The confidence map is calculated by Eq.~\eqref{eq:confidence map}. Using confidence maps, probabilistic coordinate fields (PCFs) can be obtained by integrating BCFs and confidence maps together. Technically, we binarize the confidence map (threshold taken $0.5$) and then mask the unreliable region corresponding to the BCF. The PCFs are visualized in Fig.~\ref{fig:vis_pcf}. We produce consistent positional encodings from a pair of images. The valid region encoded by the PCFs guarantees consistency in both the coordinate geometry and the content representation. Moreover, the indoor results show that our approach can identify the nonaffine region.

\subsection{Computational Complexity}
It is worth noting that our pipeline is relatively efficient because it operates on LR image pairs of size $\frac{H}{4} \times \frac{W}{4}$. We report the computational complexity and the number of PCF-Net parameters in Table~\ref{tab:compute complexity}. PCF-Net adopts an efficient network structure for probabilistic parameter prediction (See Appendix~B).

\begin{table}\footnotesize
    \centering
    \renewcommand{\arraystretch}{1.1}
    \addtolength{\tabcolsep}{1.2pt}
    \caption{Computational complexity and parameters of PCF-Net. $K$ refers to the kernel size}
    \begin{tabular}{@{}lcc@{}}
        \toprule
        Module  &FLOPs ($\times \frac{HW}{16}$) & \#Params\\ \midrule
        Compute BCFs &5 &0\\
        Coordinate encoder &35$\times 10^{3}$ $\cdot K^2$ &35$\times 10^{3}$ $\cdot K^2$\\
        Compute correlation map &20$\times 10^{3}$ $\cdot K^2$ &0\\
        Correlation decoder &1.6$\times 10^{3}$ $\cdot K^2$ &1.6$\times 10^{3}$ $\cdot K^2$\\
        Parameter predictor &2.1$\times 10^{3}$ $\cdot K^2$ &2.1$\times 10^{3}$ $\cdot K^2$\\ \midrule
        Total &58.7$\times 10^{3}$ $\cdot K^2$+5 &38.7$\times 10^{3}$ $\cdot K^2$\\
        
        \bottomrule
    \end{tabular}
\label{tab:compute complexity}
\end{table}

\section{Practical Use of PCFs} \label{sec:strategies}

Here, we show how to apply PCFs to correspondence problems. We first introduce how to build multiple coordinate systems that are used to increase the robustness of PCFs. Furthermore, considering that different correspondence tasks use different feature descriptors (\eg, different feature dimensions), multiple PCFs need different embedding strategies so that PCFs can be compatible with different descriptors. Therefore, we present possible positional encoding strategies in three typical correspondence tasks.



\subsection{Multiple Coordinate Systems}
\begin{figure}[!t]
\centering
\includegraphics[width=0.49\textwidth]{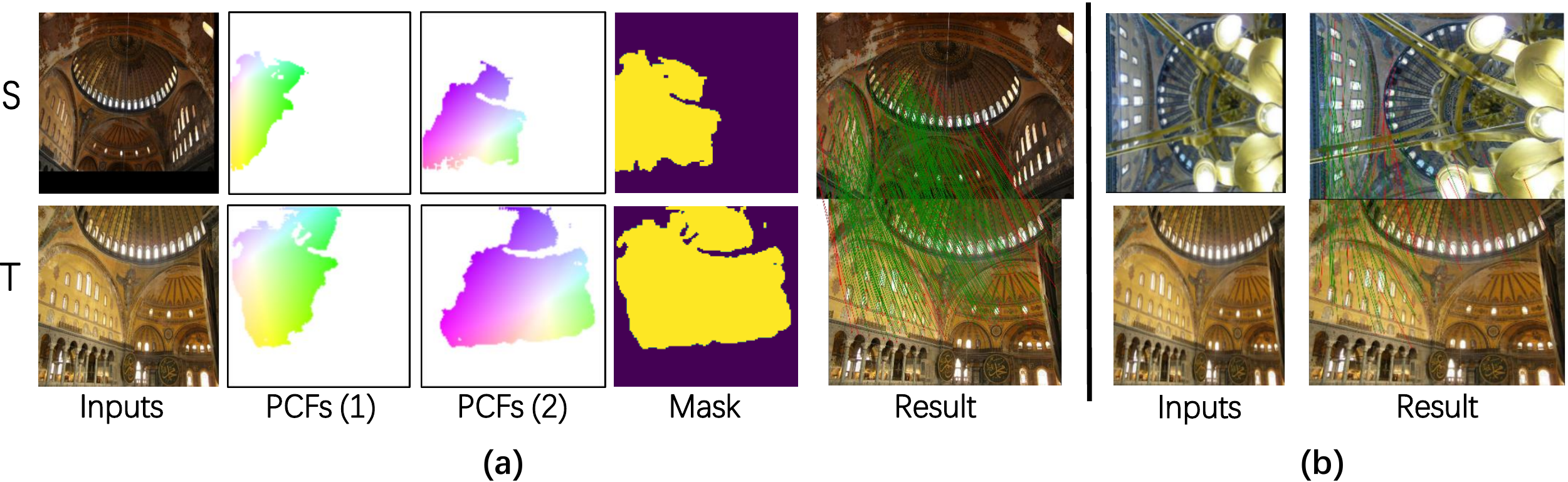}
\caption{\textbf{Visualizations of the multiple BCSs and extreme failure case.} (a) The matching result of the image pair with two different BCSs. Each BCS generates a pair of PCFs. (b) The matching result of the extreme perspective change without PCFs. The correct matches are marked in {\color{green}{green}}, and the mismatches are shown in {\color{red}{red}}.}
\label{pic:add_pic1}
\end{figure}

\begin{figure}[!t]
    \includegraphics[width=0.48\textwidth]{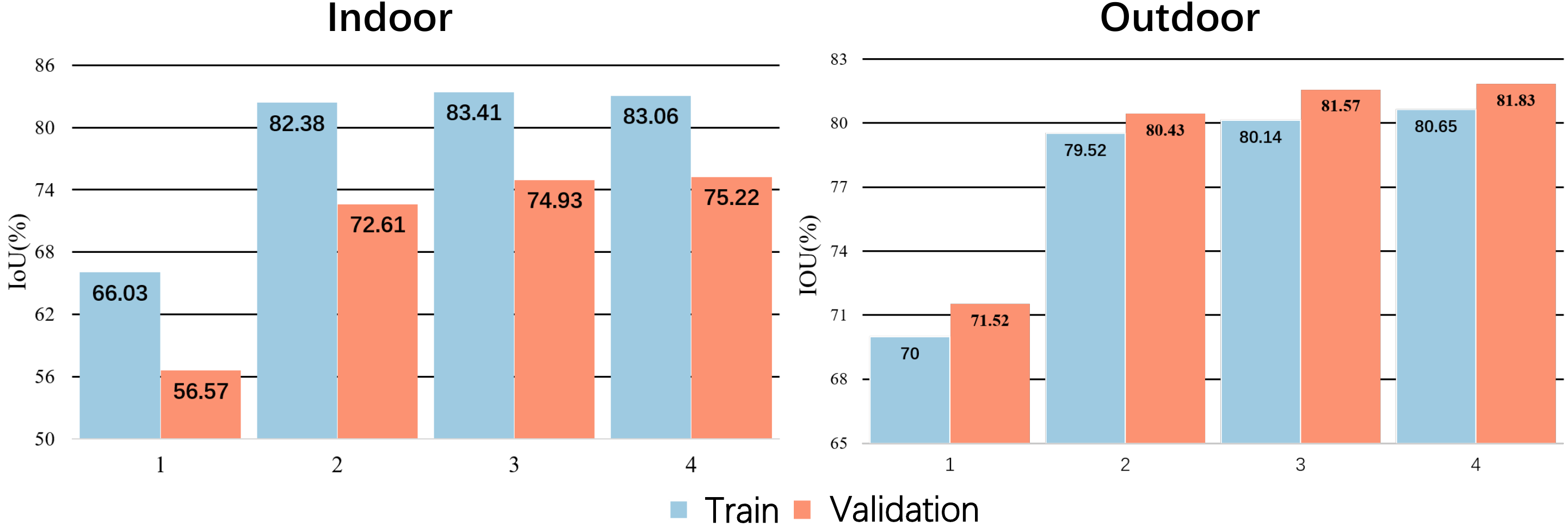}
    \caption{\textbf{Indoor and outdoor IoU evaluation.} We tested up to four different sets of BCSs.}
\label{fig:IOU}
\end{figure}

As explained in Section~\ref{subsec:Barycentric Coordinate Field}, it is difficult to encode the correct coordinates for the entire target image with a single pair of BCS. A benefit of our approach is that it allows one to obtain multiple coordinate fields from multiple BCSs. Specifically, we construct a new coordinate system outside the predicted reliable regions. However, for each additional coordinate system, the inference time increases by approximately $25$ ms. Therefore, it is important to know how many additional rewards an additional BCS pair can provide. To this end, we tested the Intersection-over-Union (IoU) metric between the union of different reliable regions and the ground-truth flow map on two large-scale datasets (MegaDepth and ScanNet),  which evaluates the additional contribution of multiple coordinate systems (Fig.~\ref{fig:IOU}). First, we find that two different pairs of BCS are sufficient to encode the entire target image. In the case of reselecting BCSs, we mask the region around the initial origin with a radius of $(K-1)/2$ and repeat the above steps to build other BCSs. Moreover, the multiple BCSs encourage reliable coordinate encoding for fitting nonrigid transforms with large relative pose, as shown in Fig.~\ref{pic:add_pic1}(a). 

\begin{figure}[!t]
    \includegraphics[width=0.48\textwidth]{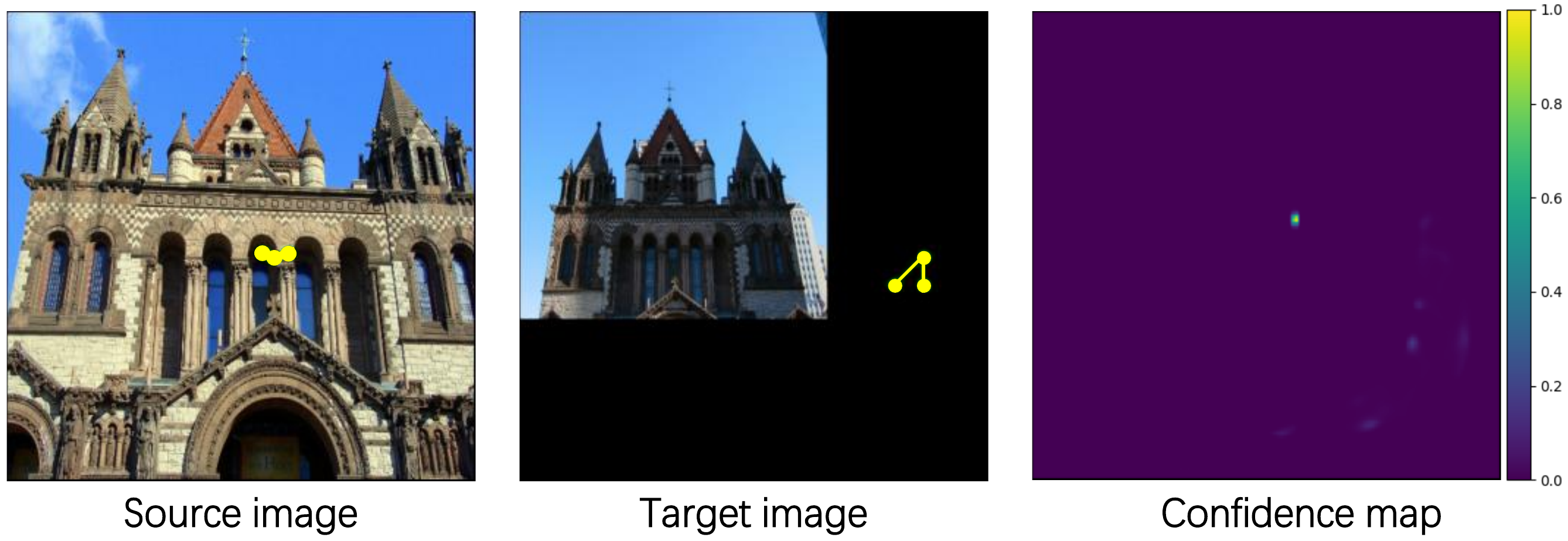}
    \caption{\textbf{Visualization of the confidence map with incorrect coordinate systems.} If the BCSs in the source and target images are incorrectly matched due to the inaccurate flow, our approach generates an almost-zero confidence map, suggesting the coordinate system should be re-built.}
\label{fig:failure case}
\end{figure}

Note that the location of the BCS origin is selected on the basis of the distribution of the estimated flow map (Section~\ref{sec:select_origin}). If the position of a coordinate system corresponds to an incorrect flow estimate, our confidence map can help to detect and reselect the coordinate system position in time (Fig.~\ref{fig:failure case}). For the failure case, we re-select the origin no more than $5$ times. If the origin is still null (\eg, the extreme perspective changes shown in Fig.~\ref{pic:add_pic1}(b), our approach returns an all-zero confidence map and the original Cartesian coordinates. Note that an all-zero confidence map is theoretically harmless to the matcher network according to our design strategy.

\subsection{Sparse Feature Matching} \label{detail:sparse}
For sparse feature matching, we investigate two different positional encoding strategies that integrate PCFs with feature descriptors: an MLP-based strategy and an attention-based one. Let $X^s \in \mathbb{R}^{N \times 2}$ and $X^t \in \mathbb{R}^{L \times 2}$ denote the \textit{sparse} normalized BCF with the zero score of the source and target image, where $N$ and $L$ represent the number of key points in the source and target image, respectively. Let $\bm{m}^s\in \mathbb{R}^{N \times 1}$ and $\bm{m}^t\in \mathbb{R}^{L \times 1}$ be the corresponding confidence values.

\vspace{5pt} \noindent\textbf{MLP-based Positional Encoding.} To use only reliable encoded coordinates, we first clip $X^{s/t}$ according to the corresponding confidence values, i.e.,
\begin{equation}\small
\hat{\bm{x}}_{i}^{s/t} =
\begin{cases}
\bm{x}_{i}^{s/t}, &\,\text{if}\,\,m_{i}^{s/t} \geq 0.5\\
\big [\max\{ X^{s/t}(:,1) \}, \max\{ X^{s/t}(:,2) \} \big ], &\,\text{otherwise}
\end{cases}\,,
\end{equation}
where $\bm{x}_{i}^{s/t} \in X^{s/t}$ denotes the barycentric coordinate of the $i^\text{th}$ keypoint. Following~\cite{sarlin2020superglue}, we embed the coordinate $\hat{\bm{x}}_i^{s/t}$ into a high-dimensional vector with an MLP. Instead of processing $\hat{X}^s$ and $\hat{X}^t$ individually~\cite{sarlin2020superglue}, we concatenate both and process them simultaneously with the same batch normalization layers.

\vspace{5pt} 
\noindent\textbf{Attention-based Positional Encoding.}
Compared to the MLP-based encoding strategy, we further develop an attention-based method with the confidence mask to capture contextual consistency. Practically, inspired by~\cite{vaswani2017attention}, we apply a transformer without positional encoding to encode $X^s$ and $X^t$ and generate $F^s \in \mathbb{R}^{N \times d}$ and $F^t \in \mathbb{R}^{L \times d}$, such that $(F^{s}, F^{t}) =\text{MLP}(X^s, X^t)$, $F^{s} = F^{s} + G_\text{self/cross}(F^{s}, F^{s/t}, \bm{m}^s, \bm{m}^{s/t})\,,$ where $G_\text{self}$ and $G_\text{cross}$ represent self-attention \cite{cho2021cats} and cross-attention with multiple heads $H = 4$, respectively, and $d$ denotes the dimension of the keypoint descriptor. In contrast to the conventional attention mechanism, we introduce the confidence maps $\bm{m}_Q$ and $\bm{m}_K$ into the formulation given by
\begin{equation}\small
    {\tt Atten}(Q,K,V,\bm{m}_Q,\bm{m}_K)={\tt softmax}\left(\frac{({\bm{m}_Q \bm{m}_K^T}) \cdot{QK}^T}{\sqrt{d_k}}\right){V} \,,
\end{equation}
where $Q$, $K$, and $V$ refer to the query, key, and value with $d$-dimension in the Transformer~\cite{vaswani2017attention}, respectively. $\bm{m}_Q$ and $\bm{m}_K$ denote the confidence maps of the query and the key. They can be the same $\bm{m}^s/\bm{m}^t$ in self-attention or $\bm{m}^s/\bm{m}^t$ and $\bm{m}^t/\bm{m}^s$ in cross-attention. Benefiting from the transformer without distance constraints, the encoded coordinate features integrate contextual and geometric positional information.

\vspace{5pt} 
\noindent\textbf{Training Details.}
We detected up to $2$ K key points for all images. The baseline is the state-of-the-art method SuperGlue~\cite{sarlin2020superglue}. We use a batch size of $20$ and train for $300$ K iterations. We use Adam optimizer~\cite{adam} with a constant leaning rate of $10^{-4}$ for the first $200/100/50$K iterations, followed by an exponential decay of $0.999998/0.999992/0.999992$ until iteration $900$ K. We perform $100$ Sinkhorn iterations and use a confidence threshold (we choose 0.2) to retain some matches from the soft assignment. For integrating sparse PCFs, we replaced the original input coordinate vectors of SuperGlue with our different encoding strategies (MLP-based/Attention-based positional encoding).

\subsection{Dense Image Registration} \label{detail:dense}
In dense image registration, we can directly embed PCFs into feature maps.
Let $C^s \in \mathbb{R}^{H \times W \times 2}$ and $C^t  \in \mathbb{R}^{H \times W \times 2} $ denote the \textit{dense} normalized BCF with zero score of the source and target image, respectively. Let $M^s\in \mathbb{R}^{H \times W \times 1}$ and $M^t\in \mathbb{R}^{H \times W \times 1}$ be the corresponding confidence maps. We choose reliable regions of BCFs with confidence maps such that
$\hat{C}_{ij} = \mathds{1}(M_{ij} - 0.5) C_{ij} \,,$
where $C_{ij}$ denotes the barycentric coordinate at the spatial location $(i, j)$, and $\hat{C}_{ij}$ is the masked coordinate at the same location. Once the PCFs $\hat{C}^s$ and $\hat{C}^t$ are obtained, we can embed them into feature maps with a few strided convolutions. Note that the input of these convolutional layers includes two different sets of BCS, resulting in $\hat{C}^{s/t} \in \mathbb{R}^{b \times 4 \times H \times W }$, where $b$ is the batch size.

\vspace{5pt} 
\noindent\textbf{Training Details.}
We choose the state-of-the-art method LoFTR~\cite{sun2021loftr} as our baseline. For the indoor dataset, the input resolution of the image pair is set to $640\times 480$. For the ourdoor dataset, the images are resized so that their longer dimensions are equal to $840$ for training and $1200$ for validation. Due to resource constraints, we use a batch size of $8$/$12$ for indoor/outdoor datasets, respectively, instead of $64$. We also reduced the original $30$ training epochs to $15$ epochs. The model is trained using Adam optimizer with an initial learning rate of $10^{-3}$. We follow the default hyperparameter settings as in LoFTR.
To integrate coordinate representations, we embed the PCFs through a series of convolutional layers illustrated above. Note that we need to resize the obtained BCFs and confidence map to fit the input resolution requirement of LoFTR.

\subsection{Consistency Filtering} \label{detail:consistency filtering}
The standard input $\mathcal{J} \in \mathbb{R}^{N \times 4}$ of consistency filtering is a concatenation of four-dimensional coordinate vectors representing $N$ candidate correspondences. Similarly, we can obtain a four-dimensional barycentric coordinate vector $\mathcal{J}_i = [\bm{x}_i^s, \bm{x}_i^t]$ for each correspondence as an alternative input, where $\bm{x}_i^s \in X^s$ and $\bm{x}_i^t \in X^t$, $i=1,...,N$. $X^s \in \mathbb{R}^{N \times 2}$ and $X^t \in \mathbb{R}^{N \times 2}$ denote the \textit{sparse} zero-score normalized BCF of the source and target image, respectively. Then, we encode the prior geometric consistency into a flag vector such that
\begin{equation}
    \tau_i = m_i^s m_i^t \cdot \frac{3\exp{(-h)} - 1}{ 1 + \exp{(-h)} } \,,
\end{equation}
where the confidence values $m_i^s$ and $m_i^t$ correspond to $\bm{x}_i^s$ and $\bm{x}_i^t$, respectively, and the coordinate distance $h = {||\bm{x}_i^s - \bm{x}_i^t||}_2$. 
In this way, reliable correspondences receive positive flags and unreliable ones receive negative flags. Moreover, the uncertain correspondences have 
almost zero values. Due to the two different pairs of BCS implemented in our experiments, the output is
$\mathcal{T} \in \mathbb{R}^{N \times 2}$. Eventually, we concatenate it with the original input $\mathcal{J}$ as the final input $\hat{\mathcal{J}} \in \mathbb{R}^{N \times 6}$, \ie, $\hat{\mathcal{J}}_i = [\bm{x}_i^s, \bm{x}_i^t, \tau_i^1, \tau_i^2]$, where $\mathcal{T}_i = [\tau_i^1, \tau_i^2]$, and $\tau_i^j$ indicates the $i^{\text{th}}$ row and the $j^{\text{th}}$ column of $\mathcal{T}$.

\vspace{5pt} 
\noindent\textbf{Training Details.}
Our baseline is chosen as the state-of-the-art outlier rejection network OANet~\cite{zhang2019learning}. 
We use Adam solver with a learning rate of $10^{-4}$ and batch size $32$. The weight $\alpha$ is 0 during the first $20$ K iterations and then $0.1$ in the rest $480$ K iterations as in~\cite{yi2018learning}. We detect $2$ K key points for each image evaluated.
For the rest, we used the default hyperparameters in the original implementation. The initial matching set is generated with the mutual nearest neighbor check (MNN) and the ratio test (RT)~\cite{lowe2004distinctive}. Following the above strategy, we re-encode the PCFs to a $2$D vector and concatenate it with the original $4$D input vector.

\section{Experiments}
\label{sec:exp}
We integrate our approach into a lightweight implementation of an optical flow network, GLU-Net~\cite{truong2020glu}, and perform extensive experiments on multiple matching datasets. We show that PCF-Net can work with various descriptors to achieve state-of-the-art performance and that our approach can be applied to different correspondence problems, including sparse feature matching, dense image registration, and consistency filtering. We also identify the specificity of our constrained Gaussian mixture model and will thus highlight the differences between our model and PDCNet~\cite{truong2021learning}.
Moreover, we demonstrate the potential of our approach in texture transfer and multi-homography classification.

\subsection{Datasets}
Here we introduce the details of all datasets used, including data sampling and splitting strategies.

\vspace{5pt} \noindent\textbf{Synthetic Dataset.} We use a mixture of synthetic data for PCF-Net training with a total of $40$ K images, which combines datasets from DPED~\cite{ignatov2017dslr}, CityScapes~\cite{cordts2016cityscapes}, and ADE-20K~\cite{zhou2019semantic}. Following~\cite{truong2020glu}, training image pairs are generated by applying random warps and small local perturbations to the original images. Meanwhile, for better compatibility with real scenes, we further augment the synthetic data with a random moving object from the MS-COCO~\cite{lin2014microsoft} dataset as implemented in~\cite{truong2021learning}.

\vspace{5pt} \noindent\textbf{MegaDepth.} MegaDepth~\cite{2018MegaDepth} is a large-scale outdoor dataset consisting of $196$ scenes, which are reconstructed from one million images on the Internet using COLMAP~\cite{schonberger2016structure}. We generate ground truth correspondences by projecting all points of the source image with depth information onto the target image, using the intrinsic and extrinsic camera parameters provided by D2-Net~\cite{Dusmanu2019CVPR}. A depth check of the source depth map is also conducted to remove irrelevant pixels such as sky and pedestrians. For PCF-Net, we use $150$ scenes and sample up to $58$ K training pairs with an overlap ratio of at least $30\%$ in the sparse SfM point cloud. Furthermore, we sampled $1800$ validation data from $25$ different scenes. For dense image registration, we follow~\cite{tyszkiewicz2020disk} to only use the scenes of ``Sacre Coeur'' and ``St. Peter's Square'' for validation. Training and testing indices are provided by~\cite{sun2021loftr}. 

\vspace{5pt} \noindent\textbf{ScanNet.} ScanNet~\cite{dai2017scannet} is a large-scale indoor dataset composed of monocular sequences with ground-truth poses and depth images. It also has well-defined training, validation, and testing splits. 
We generate ground-truth correspondences referring to the procedure used on the MegaDepth dataset. Following the dataset indices provided by~\cite{sarlin2020superglue}, we select $230$ M training pairs and $1500$ testing pairs, discarding pairs with small or large overlaps.

\vspace{5pt} \noindent\textbf{YFCC100M.} Yahoo's YFCC100M dataset~\cite{thomee2016yfcc100m} contains $100$ M internet photos, and \cite{heinly2015reconstructing} later generates $72$ $3$D reconstructions of tourist landmarks from a subset of the collections. Following~\cite{zhang2019learning}, we select $68$ scenes for training and $4$ scenes for testing. In each scene, image pairs with overlap beyond $50\%$ are included in the data set, resulting in pairs of $250$K training and $4$K testing. Due to the lack of depth maps, ground-truth correspondences are supervised by the symmetric epipolar distance ($< 1e^{-4}$).

\vspace{5pt} \noindent\textbf{PhotoTourism.} PhotoTourism~\cite{jin2020image} is a subset of the YFCC100M dataset~\cite{thomee2016yfcc100m} with $15$ scenes and has ground-truth poses and sparse 3D models obtained from COLMAP. We select $12$ scenes for training and the rest for testing. In each scene, we generate image pairs by finding the top $10$ images for each image according to the number of common points, resulting in image pairs of $230$ K training and $6$ K testing.

\vspace{5pt} \noindent\textbf{SUN3D.} The SUN3D dataset~\cite{xiao2013sun3d} is an indoor RGBD video dataset with camera poses computed by generalized bundle adjustment. Following~\cite{zhang2019learning}, we split the dataset into sequences, with $239$ for training and $15$ for testing. We subsample each video every $10$ frames and select image pairs with an overlap beyond $35\%$. Finally, the $1$ M training and $1500$ testing pairs are selected.

\subsection{Protocols and Implementation Details of PCF-Net}
\label{sec:details}

\noindent\textbf{Training Details.}
To train the PCF-Net, we use both synthetic and real data. Synthetic data include DPED~\cite{ignatov2017dslr}, CityScapes~\cite{cordts2016cityscapes}, ADE-20K~\cite{zhou2019semantic}, and COCO~\cite{lin2014microsoft}. Real data uses MegaDepth~\cite{2018MegaDepth} and ScanNet~\cite{dai2017scannet} datasets. 

We adopt VGG-16~\cite{VGG16} as the feature extractor backbone and GLU-Net~\cite{truong2020glu} as the flow network. The input image pairs of VGG-16 are cropped to $520 \times 520$ during training. Our PCF-Net cascades separately after the last two layers of the GLU-Net (which corresponds to resolutions of $1/4$ and $1/8$). The parameters in Eq.~\eqref{eq:constrained sigma} are fixed to $\delta_+$ = $1$, $\delta_-$ = $11$, and $\Delta\delta$ = $2$. $\gamma$ in Eq.~\eqref{eq:distance map} is set to $0.03$. The radius $R$ in Eq.~\eqref{eq:confidence map} is set to $1$.

Training has two stages. In the first stage, we employ the VGG-16 network pre-trained on ImageNet and the GLU-Net pre-trained on DPED-CityScape-ADE~\cite{truong2020glu}, and PCF-Net is trained on synthetic data. Note that during the first training stage, the feature and flow backbones are frozen. In the second stage, we train our approach on a combination of sparse data from real scenes and dense synthetic data. In addition, we fine-tune the two backbones in this stage.

During the first stage of training on unique synthetic data, we train for $130$ epochs, with a batch size of $16$. The learning rate is initially set to $10^{-4}$ and halved after the $70$ and $110$ epochs. When fine-tuning the composition of the real-scene dataset and synthetic dataset, the batch size is reduced to $8$, and we train for the $150$ epochs. The initial learning rate is fixed at $5\times10^{-5}$ and halved after $90$ and $130$ epochs. Our approach is implemented using PyTorch~\cite{pytorch} and our networks are trained using the Adam optimizer~\cite{adam} with a weight decay of $0.0004$.

\vspace{5pt}
\noindent\textbf{Evaluation Metrics.} To evaluate the performance of correspondence selection, we report the Precision (P), Recall (R), and F-measure (F) as in~\cite{zhang2019learning,sarlin2020superglue}. We use two types of precision for different tasks. For consistency filtering~\cite{zhang2019learning}, P${_{epi}}$ represents the epipolar distance of the correspondences below ${10^{-4}}$. For sparse feature matching~\cite{sarlin2020superglue}, P$_{proj}$ indicates the reprojection error of correspondences lower than $5$ pixels. To further measure the accuracy of pose estimation, we report the AUC of pose error at thresholds of $5^\circ$, $10^\circ$, and $20^\circ$. 
Note that the AUC metric adopts the approximate AUC as in~\cite{zhang2019learning,sun2021loftr}.
To recover the pose of the camera, we calculate the essential matrix with ${\tt findEssentialMat}$ (the threshold is set to $0.001$) implemented by OpenCV and RANSAC, followed by ${\tt recoverPose}$.

\subsection{Results on Correspondence Tasks} \label{subsec:superlue}

\begin{table}[t] \scriptsize
\centering
\addtolength{\tabcolsep}{0pt}
\renewcommand{\arraystretch}{1.2}
\caption{Ablation study of designing choices in PCF-Net. Upper sub-table uses the RootSIFT descriptor and the lower subtable uses the SuperPoint descriptor. CC: Cartesian coordinates; BCF: barycentric coordinate field; CM: confidence map estimated by PCF-Net; PE: positional encoding that integrates BCF with feature descriptors; MLP: multilayer perceptron; Atten.: masked attentional mechanisms in Transformer~\cite{vaswani2017attention}. The best performance is in \textbf{boldface}, and the second best is \underline{underlined}}

\begin{NiceTabular}{@{}ccc c ccccc@{}}[colortbl-like]
\toprule
  \multirow{2}{*}{CC} & \multirow{2}{*}{BCF} & \multirow{2}{*}{CM} & \multirow{2}{*}{PE}  & \multicolumn{3}{c}{AUC}   & \multirow{2}{*}{P${_{\tt proj}}$} & \multirow{2}{*}{R} \\ \cmidrule(lr){5-7}
  ~  & ~ & ~ & ~  &\makecell*[c]{@${5^{\circ}}$}  & @${10^{\circ}}$ & @${20^{\circ}}$ &  ~  & ~                     \\ \hline
\rowcolor{gray!5} \checkmark & ~ & ~ & -  &  58.83          &  66.30          &  74.74          &  73.50             &  50.53                \\
\rowcolor{gray!20} ~ & \checkmark & ~ & MLP &  57.42          &  66.51          &  74.20          &  71.62             &  53.32                 \\
\rowcolor{gray!5} \checkmark & \checkmark & ~ & MLP &  57.67          &  66.74          &  75.89          &  72.38             & {\underline{58.63}}   \\
\rowcolor{gray!20} \checkmark & \checkmark & \checkmark & MLP & {\underline{59.35}}    & {\underline{67.35}}    & {\underline{76.47}}    & {\underline{75.17}}   &  56.76   \\
\rowcolor{gray!5} \checkmark & \checkmark & \checkmark & Atten.  & \textbf{60.17} & \textbf{68.16} & \textbf{77.33} & \textbf{77.21}    & \textbf{62.91}   \\  \midrule

\rowcolor{gray!5} \checkmark & ~ & ~ & -  & 65.36   & 73.71   & 81.64   & 76.92  & 75.75  \\
\rowcolor{gray!20} ~ & \checkmark & ~ & MLP & 65.84    &73.54      &81.37 &77.15     &74.83     \\
\rowcolor{gray!5} \checkmark & \checkmark & ~ & MLP & 65.88   & 74.09      & 82.13   &77.47  &76.36   \\
\rowcolor{gray!20} \checkmark & \checkmark & \checkmark & MLP & \underline{66.05}  & \textbf{74.50}  & \underline{82.64}  & \underline{78.71}  &\underline{76.59}   \\
\rowcolor{gray!5} \checkmark & \checkmark & \checkmark & Atten.   & \textbf{66.13} & \underline{74.44} & \textbf{82.81} & \textbf{78.92}  &\textbf{76.83}  \\ 

\bottomrule
\end{NiceTabular}
\label{table:ablation study in superglue}
\end{table}

\begin{table}[t]\scriptsize
\centering
\addtolength{\tabcolsep}{-0.5pt}
\renewcommand{\arraystretch}{1.1}
\caption{Comparison with SuperGlue on the PhotoTourism dataset. We report the AUC of pose errors, precision (P$_{\tt proj}$), recall (R), and F-measure (F), all in percentage. P$_{\tt proj}$ denotes the mean matching precision measured by the projection distance error. 
SuperGlue$^{*}$ indicates the results of the officially released model pretrained on MegaDepth dataset~\cite{2018MegaDepth}.
Best performance is in \textbf{boldface}
}
\begin{NiceTabular}{@{}llcccccc@{}}[colortbl-like]
\toprule
\multirow{2}{*}{Descriptor} & \multirow{2}{*}{Matcher} & \multicolumn{3}{c}{AUC}                             & \multirow{2}{*}{P$_{\tt proj}$} & \multirow{2}{*}{R} & \multirow{2}{*}{F} \\ \cline{3-5}
                            &                   &\makecell*[c]{ @${5^{\circ}}$}  & @${10^{\circ}}$ & @${20^{\circ}}$ &                    &                    &                    \\ \hline
\rowcolor{gray!5} & SuperGlue         &  58.83          &  66.30          &  74.74          &  73.50             &  50.53             &  58.53             \\
\rowcolor{gray!5} \multirow{-2}{*}{RootSIFT}  & Ours              & \textbf{60.17} & \textbf{68.16} & \textbf{77.33} & \textbf{77.21}    & \textbf{62.91}    & \textbf{68.31}    \\ 
\rowcolor{gray!20}    & SuperGlue         &  32.56          &  41.59          &  52.32          &  54.45             &  27.41             &  34.12             \\
\rowcolor{gray!20}    \multirow{-2}{*}{HardNet}   & Ours              & \textbf{41.22} & \textbf{52.04} & \textbf{63.86} & \textbf{62.57}    & \textbf{41.52}    & \textbf{48.31}    \\
\rowcolor{gray!5}  & SuperGlue$^{*}$  &65.87 &\textbf{76.97} &\textbf{85.23} &\textbf{80.54} &74.48 &\textbf{78.13}        \\  
\rowcolor{gray!5}  & SuperGlue         &  65.36          &  73.71          &  81.64          &  76.92             &  75.75             &  76.05             \\       
\rowcolor{gray!5}  \multirow{-3}{*}{SuperPoint}  & Ours              & \textbf{66.13} & 74.44 & 82.81 & 78.92    & \textbf{76.83}    & 77.65    \\ \hline
\rowcolor{gray!20}  \multirow{0.5}{*}{Detector-free}  & {LoFTR-OT}    & {69.19}   & {78.23}   & {84.95}   &{57.22}        &  -    &   -         \\
\bottomrule
\end{NiceTabular}
\label{table:superglue-outdoor}
\end{table}

\begin{table}[t]\scriptsize
\centering
\addtolength{\tabcolsep}{0pt}
\renewcommand{\arraystretch}{1.1}
\caption{Comparison with SuperGlue on the SUN3D dataset. Our approach improves 
SuperGlue with all descriptors, especially on the local-context-only descriptor HardNet. SuperGlue$^{*}$ indicates the results of the officially released model pretrained on ScanNet dataset~\cite{dai2017scannet}. Best performance is in \textbf{boldface}}
\begin{NiceTabular}{@{}llcccccc@{}}[colortbl-like]
\toprule
\multirow{2}{*}{Descriptor} & \multirow{2}{*}{Matcher} & \multicolumn{3}{c}{AUC}                             & \multirow{2}{*}{P$_{\tt proj}$} & \multirow{2}{*}{R} & \multirow{2}{*}{F} \\  \cline{3-5}
                            &                   &\makecell*[c]{ @${5^{\circ}}$}  & @${10^{\circ}}$ & @${20^{\circ}}$ &                    &                    &                    \\  \hline
\rowcolor{gray!5}   & SuperGlue         &   12.18          &   19.56          &   29.60          & \textbf{52.05}            &   25.82             &   23.57             \\
\rowcolor{gray!5} \multirow{-2}{*}{RootSIFT}   & Ours              & \textbf{12.56} & \textbf{20.83} & \textbf{32.47} & 51.81    & \textbf{31.18}    & \textbf{37.64}    \\ 
\rowcolor{gray!20}   & SuperGlue         &   12.41          &   20.29          &   32.02          &   49.21             &   31.59             &   37.52             \\
\rowcolor{gray!20}   \multirow{-2}{*}{HardNet}        & Ours              & \textbf{14.75} & \textbf{23.21} & \textbf{35.24} & \textbf{51.18}    & \textbf{35.46}    & \textbf{40.96}    \\

\rowcolor{gray!5}  & {SuperGlue$^{*}$}  &{13.52} &{21.86} &{32.67} &{50.50} &{31.34} &{37.35}    \\ 
\rowcolor{gray!5}  & SuperGlue         &   14.38          &   23.29          &   35.13          &   51.19             &   31.73             &   37.96             \\ 
\rowcolor{gray!5} \multirow{-3}{*}{SuperPoint} & Ours              & \textbf{17.60} & \textbf{27.61} & \textbf{38.27} & \textbf{52.71}    & \textbf{36.37}    & \textbf{43.64}      \\ \hline
\rowcolor{gray!20}  \multirow{0.5}{*}{{Detector-free}}  & {LoFTR-OT}  &{18.57}  &{29.13}  &{39.48}   &{38.45}     &  -   &  -      \\
\bottomrule
\end{NiceTabular}
\label{table:superglue-indoor}
\end{table}

\begin{figure}[!t]
\centering  
\includegraphics[width=0.49\textwidth]{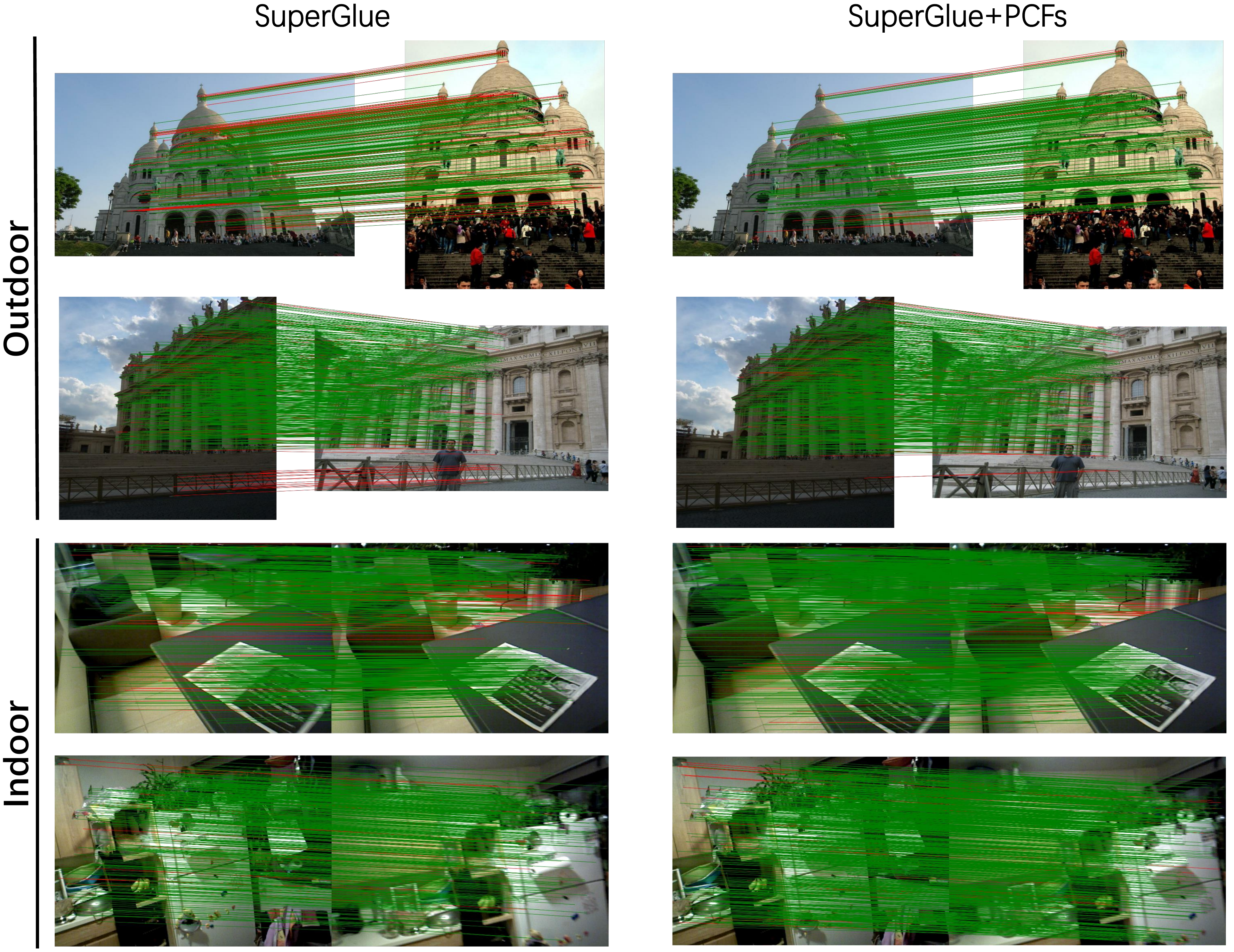}
\caption{\textbf{Visualization of SuperGlue~\cite{sarlin2020superglue} with RootSIFT~\cite{arandjelovic2012three} on the PhotoTourism and SUN3D datasets.} Correct matches are marked in {\color{green}{green}}, and mismatches are shown in {\color{red}{red}}. Note that PCFs consistently improve the matching performance of SuperGlue in all situations.} 
\label{fig:superglue}
\end{figure}

\subsubsection{Sparse Feature Matching}
Sparse feature matching is often affected by the quality of the descriptors. In particular, coordinate representations are largely overlooked in this field. To demonstrate the superiority of PCFs for sparse feature matching, we built our baseline using the state-of-the-art network SuperGlue~\cite{sarlin2020superglue} and three representative descriptors: RootSIFT~\cite{arandjelovic2012three}, HardNet~\cite{mishchuk2017working}, and SuperPoint~\cite{detone2018superpoint}. We replace the original SuperGlue input coordinate vectors with our sparse PCFs (see Section~\ref{detail:sparse}). We report the performance on the PhotoTourism~\cite{jin2020image} and SUN3D~\cite{xiao2013sun3d} datasets. Note that SuperGlue~\cite{sarlin2020superglue} did not provide the official training code. Therefore, we have tried our best to reproduce the results of the original article by consulting the authors for a fair comparison.

\vspace{5pt}
\noindent\textbf{Ablation Study.} To understand the importance of PCF-Net and explore the appropriate implementation of PCFs, we first conduct an ablation study on the PhotoTourism dataset with RootSIFT and SuperPoint descriptors. The results are shown in Table~\ref{table:ablation study in superglue}: 
\romannumeral1) The naive BCF encoded by PCF-Net consistently improves recall; 
\romannumeral2) The combination of Cartesian coordinates and barycentric coordinates performs better than independently using each of them; 
\romannumeral3) The confidence map contributes to a significant performance improvement ($+2.6\%$ at AUC@$20^\circ$ and $+12\%$ at recall for RootSIFT), which implies that the reliability of geometric coherence appears to be more important than uninformed geometric invariance; 
\romannumeral4) Compared with conventional positional encoding by MLP (see Section~\ref{detail:sparse}), the attentional mask works better in encoding geometric positional information. 
In the following experiments, we adopt the last row of the sub-tables in Table~\ref{table:ablation study in superglue} as our baseline (CC+BCF+CM+Atten.).

\vspace{5pt}
\noindent\textbf{Results.}
Outdoor results are reported in Table~\ref{table:superglue-outdoor}. Our approach (PCF-Net + SuperGlue) improves pose estimation and matching accuracy using three different descriptors, especially when HardNet is used. Due to the patch-based characteristic of HardNet, SuperGlue can acquire only global information from keypoint coordinates.
However, the poor performance of SuperGlue with Cartesian coordinates implies that Cartesian coordinates are not useful. By integrating reliable barycentric coordinates into SuperGlue, a significant improvement is observed in all metrics ($+7.8\%$ in precision, $+14.1\%$ in recall, and $+11.5\%$ in pose estimation). 
These results clearly demonstrate the advantages of reliable coordinate representation and an appropriate encoding strategy in sparse feature matching. 

We also remark that the improvements of PCF-Net in the indoor dataset (Table~\ref{table:superglue-indoor}) are not as obvious as in outdoor scenes, because low-texture regions or repetitive patterns appear more frequently in indoor environments. But SuperGlue with PCFs still shows a certain degree of superiority over the baseline, especially in pose estimation with $+3\%$ in AUC@$20^\circ$. 
Particularly, SuperGlue with PCFs and the SuperPoint descriptor achieves competitive results with the detector-free method LoFTR~\cite{sun2021loftr}. 
Visualizations of the correspondence results are shown in Fig.~\ref{fig:superglue}. 

\begin{figure}[!t]
\centering  
\includegraphics[width=0.49\textwidth]{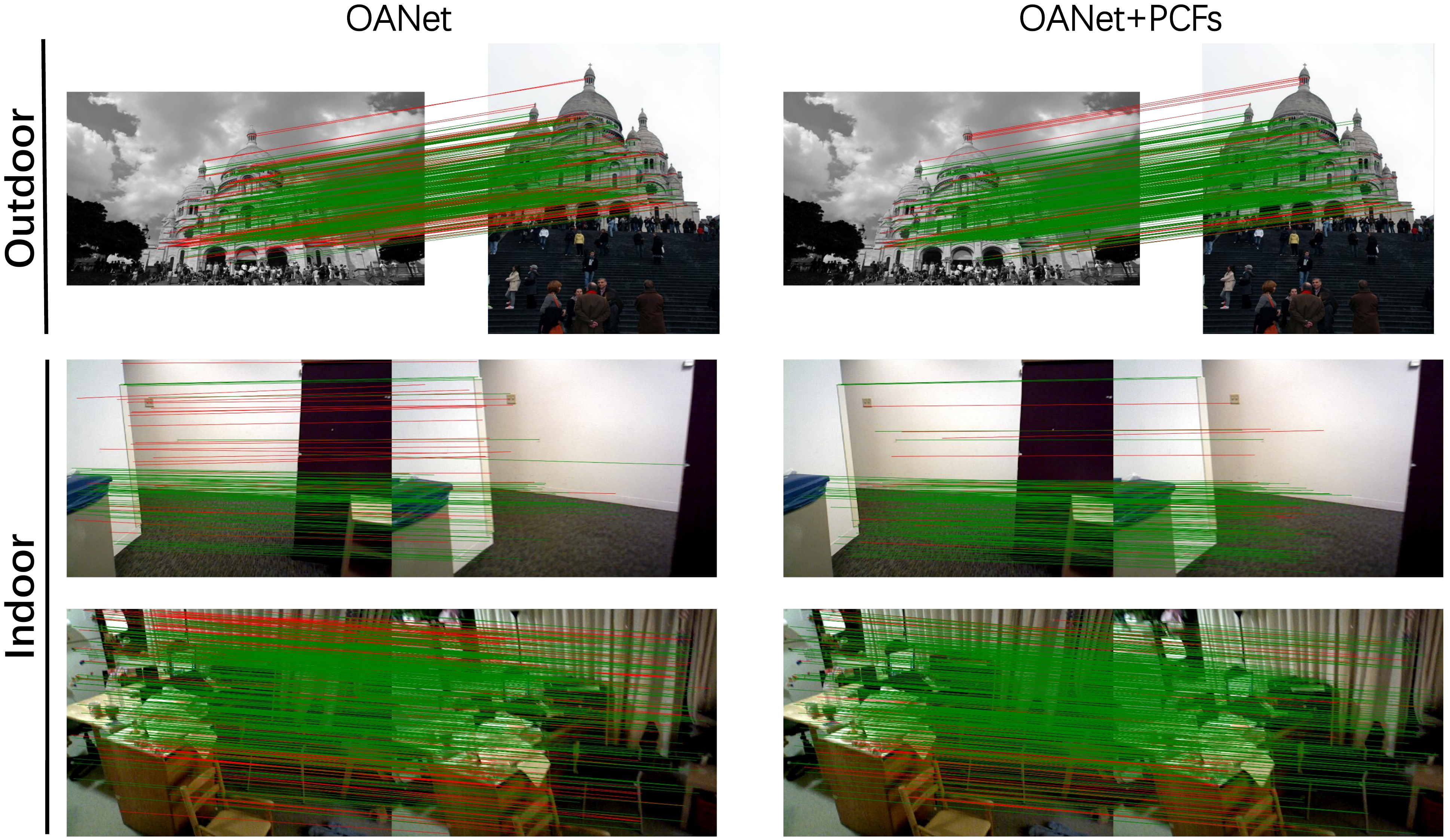}
\caption{\textbf{Visualization of OANet~\cite{zhang2019learning} with RootSIFT on YFCC100M and SUN3D datasets.} Note that PCFs consistently improve the matching performance of OANet in all situations.} 
\label{fig:oanet}
\end{figure}

\begin{table*}[!t] \footnotesize
\centering
\renewcommand{\arraystretch}{1.1}
\addtolength{\tabcolsep}{10pt}
\caption{Evaluation with LoFTR on Megadepth and ScanNet for pose estimation. OT: differentiable matching with optimal transport~\cite{sarlin2020superglue}; DS: differentiable matching with dual softmax~\cite{rocco2018neighbourhood}; w.o.~sin: the baseline without sine positional encoder. The best performance is in \textbf{boldface}} 

\begin{NiceTabular}{@{}lcccccc@{}}[colortbl-like]
\toprule
\multirow{2}{*}{Method} & \multicolumn{3}{c}{AUC -- MegaDepth~\cite{2018MegaDepth}} & \multicolumn{3}{c}{AUC -- ScanNet~\cite{dai2017scannet}} \\ \cmidrule(lr){2-4} \cmidrule(lr){5-7}
       ~   &\makecell*[c]{ @${5^{\circ}}$}  & @${10^{\circ}}$ & @${20^{\circ}}$    &\makecell*[c]{ @${5^{\circ}}$}  & @${10^{\circ}}$ & @${20^{\circ}}$ \\ \hline
\rowcolor{gray!0} DS (baseline)        & 46.92          & 63.71       & 76.65   & 19.81 &37.64 &54.27 \\
\rowcolor{gray!0} DS (baseline) + PCFs     & 48.27          & 65.05       & 77.53  & 20.86  &39.22  &56.33\\
\rowcolor{gray!0} DS (w.o. sin) + PCFs &\textbf{48.85}     &\textbf{65.24}     &\textbf{78.40}  &\textbf{21.44} &\textbf{39.88} &\textbf{57.02} \\ \hline
\rowcolor{gray!0} OT (baseline)       & 47.38    &  64.11  & 76.64  & 19.35 &37.30  & 54.24 \\
\rowcolor{gray!0} OT (baseline) + PCFs     & 49.17  & \textbf{65.62}  & 77.93  & 20.81  &39.38 & 56.59\\
\rowcolor{gray!0} OT (w.o. sin) + PCFs   &\textbf{49.81}  & 65.40 & \textbf{78.98} &\textbf{21.47} &\textbf{39.78} &\textbf{57.18} \\
\bottomrule
\end{NiceTabular}
\label{table:loftr-outdoor-indoor}
\end{table*}


\subsubsection{Dense Image Registration}
Although CNNs implicitly encode position-sensitive features, dense image registration approaches typically use $2$D extension of frequency sine functions to extract positional information. However, such positional information can be ambitious for an image pair.
Here, we further demonstrate the advantage of PCF in dense image registration. We choose the state-of-the-art detector-free network LoFTR~\cite{sun2021loftr} as our baseline. To integrate coordinate representations, we embed the PCFs into feature maps through $3$ successive stride-$2$ convolutional layers with the kernel size of $3 \times 3$ (see Section~\ref{detail:dense}). Following LoFTR, we use MegaDepth~\cite{2018MegaDepth} and ScanNet~\cite{dai2017scannet} for the evaluation.

\vspace{5pt}
\noindent\textbf{Results.}
As shown in Table~\ref{table:loftr-outdoor-indoor}, compared to the baseline in both datasets with two types of differentiable matching layers, LoFTR with PCFs achieves better performance in pose estimation (on average $+2.5\%$ in AUC). The improved results again demonstrate the importance of geometrically invariant coordinate encoding.
Interestingly, when we remove the original position encoder from the baseline, the results are improved, suggesting that inappropriate positional representation can even degrade the performance in dense image registration.

\begin{table}[!t]\scriptsize
\centering
\caption{Comparison with OANet on the YFCC100M dataset. NN denotes nearest neighbor matching with the mutual check and ratio test, and P$_{\tt epi}$ denotes the mean matching precision with the epipolar geometric distance error. The best performance is in \textbf{boldface}}
\addtolength{\tabcolsep}{-0.7pt}
\renewcommand{\arraystretch}{1.1}
\begin{NiceTabular}{@{}llcccccc@{}}[colortbl-like]
\toprule
\multirow{2}{*}{Descriptor} & \multirow{2}{*}{Method} & \multicolumn{3}{c}{AUC}                             & \multirow{2}{*}{P$_{\tt epi}$} & \multirow{2}{*}{R} & \multirow{2}{*}{F} \\ \cmidrule(r){3-5}
                            &       &\makecell*[c]{ @${5^{\circ}}$}  & @${10^{\circ}}$ & @${20^{\circ}}$ &                    &                    &                    \\ \hline
\rowcolor{gray!5}   & NN+OANet  &  54.60   &  64.57    & 74.25    &  46.10   & 79.50   & 55.20   \\
\rowcolor{gray!5} \multirow{-2}{*}{RootSIFT} & Ours      & \textbf{55.80} & \textbf{65.72} & \textbf{77.23} & \textbf{57.31}    & \textbf{86.10}    & \textbf{67.82}    \\ 
\rowcolor{gray!20}     & NN+OANet  &  56.20          &  66.72          &  76.91          &  67.83             &  87.85             &  74.90            \\
\rowcolor{gray!20} \multirow{-2}{*}{HardNet}     & Ours      & \textbf{57.63} & \textbf{67.19} & \textbf{79.93} & \textbf{70.10}    & \textbf{89.64}    & \textbf{77.86}    \\
\rowcolor{gray!5} & NN+OANet  & \textbf{41.83}    &  54.25          &  67.93          &  64.63     & \textbf{87.02 }  & 72.04              \\                     
\rowcolor{gray!5} \multirow{-2}{*}{SuperPoint} & Ours      & 41.65    & \textbf{54.76} & \textbf{68.59} & \textbf{66.79}    &  86.24    & \textbf{73.28}    \\ 
\bottomrule
\end{NiceTabular}
\label{table:oanet-outdoor}
\end{table}

\begin{table}[!t]\scriptsize
\centering
\caption{Comparison with OANet on the SUN3D dataset. The best performance is in \textbf{boldface}}
\addtolength{\tabcolsep}{-0.7pt}
\renewcommand{\arraystretch}{1.1}
\begin{NiceTabular}{@{}llcccccc@{}}[colortbl-like]
\toprule
\multirow{2}{*}{Descriptor} & \multirow{2}{*}{Method} & \multicolumn{3}{c}{AUC}                             & \multirow{2}{*}{P$_{\tt epi}$} & \multirow{2}{*}{R} & \multirow{2}{*}{F} \\ \cline{3-5}
                            &                   &\makecell*[c]{ @${5^{\circ}}$}  & @${10^{\circ}}$ & @${20^{\circ}}$ &                    &                    &                    \\ \hline
\rowcolor{gray!5} & NN+OANet         & \textbf{17.42}          &  26.82          &  39.29          &  47.87             &  84.59             &  57.48             \\
\rowcolor{gray!5}   \multirow{-2}{*}{RootSIFT}     & Ours              & 17.33  & \textbf{27.86} & \textbf{41.37} & \textbf{50.70}    & \textbf{88.22}    & \textbf{60.53}    \\ 
\rowcolor{gray!20}     & NN+OANet         &  17.23          &  26.72          &  39.32          &  56.17             &  86.57             &  65.35             \\
\rowcolor{gray!20} \multirow{-2}{*}{HardNet} & Ours              & \textbf{18.05} & \textbf{27.61} & \textbf{41.19} & \textbf{57.26}    & \textbf{89.14}    & \textbf{67.34}  \\
                         
\rowcolor{gray!5} & NN+OANet         &  19.03          & \textbf{29.32}          &  43.01          &  64.48             &  86.35   &  71.26             \\       
\rowcolor{gray!5}  \multirow{-2}{*}{SuperPoint}     & Ours              & \textbf{19.41} &  28.97 & \textbf{44.73} & \textbf{66.79}    & \textbf{86.95}    & \textbf{74.34}    \\ 
\bottomrule
\end{NiceTabular}
\label{table:oanet-indoor}
\end{table}

\subsubsection{Consistency Filtering}

The standard input for consistency filtering is an $4$-element vector that concatenates the coordinates of each correspondence. Intuitively, such a coordinate vector would lack global geometric information. Here, we explore the potential of PCFs in this task. In particular, we reencode the PCFs to a $2$D vector that characterizes the initial possibility that correspondences are inliers/outliers (see Section~\ref{detail:consistency filtering}). This $2$D vector is then concatenated into the standard coordinate input as the new representation to a consistency filter.
We choose the state-of-the-art outlier rejection network OANet~\cite{zhang2019learning} as the baseline. Following OANet, we report on performance on the Yahoo YFCC100M~\cite{thomee2016yfcc100m} and SUN3D datasets.

\vspace{5pt}
\noindent\textbf{Results.}
The results are listed in Tables~\ref{table:oanet-outdoor} and~\ref{table:oanet-indoor}. In both datasets, PCFs achieve a clear improvement in outlier rejection and pose estimation. Even without the learnable parameters for the encoded coordinates, our method still outperforms the baseline (on average $+2.3\%$ in the AUC) with all descriptors. In particular, in the setting of RooSIFT in the outdoor dataset, we observe a relative improvement of $+11.2\%$ in precision and $+6.6\%$ in recall. Our results imply that PCF-Net reliably encodes geometrically invariant coordinates between image pairs, and geometric coherence is also crucial for consistency filtering. Visualizations of the correspondence results are shown in Fig.~\ref{fig:oanet}. 
It should be noted that there are discrepancies in the results obtained by our reproduction of OANet using SuperPoint features and the results reported in Appendix of SuperGlue~\cite{sarlin2020superglue}. These discrepancies can be attributed to the following differences in the experimental settings used for SuperPoint feature extraction: \romannumeral1) We set the non-maximum suppression (nms) parameter to 3 instead of 4, and \romannumeral2) We directly extracted the SuperPoint descriptors without resizing the input images.

\subsection{Comparison of Competing Probabilistic Models} \label{subsec:experi_probablistic}

\begin{table}[!t] \footnotesize
\centering
\renewcommand{\arraystretch}{1.1}
\addtolength{\tabcolsep}{0.8pt}
\caption{Ablation study of our probabilistic model. IoU $(\%)$ performance comparison of different probabilistic model settings on two datasets. We gradually change the probabilistic model settings to explore the contribution of each step. The best performance is in \textbf{boldface}} 
\rowcolors{2}{gray!20}{gray!5}
\begin{tabular}{@{}>{\columncolor{white}[0pt][\tabcolsep]}lc>{\columncolor{white}[\tabcolsep][0pt]}c@{}}
\toprule
Settings &  \multicolumn{1}{c}{MegaDepth~\cite{2018MegaDepth}}   &  \multicolumn{1}{c}{ScanNet~\cite{dai2017scannet}}     \\ \hline
flow probabilistic model~\cite{truong2021learning}    & 14.75         & 9.63       \\
{fixed $\sigma_+$ $\Rightarrow$ learnable $\sigma_+$} &23.66  &21.84 \\
{softmax $\Rightarrow$ sigmoid} &63.71  &50.39 \\
{laplacian $\Rightarrow$ Gaussian} (ours) & \textbf{71.52} & \textbf{56.57} \\ \bottomrule
\end{tabular}
\label{table:pro_model}
\end{table}

\begin{figure}[!t]
  \centering
  \includegraphics[width=1.0\linewidth]{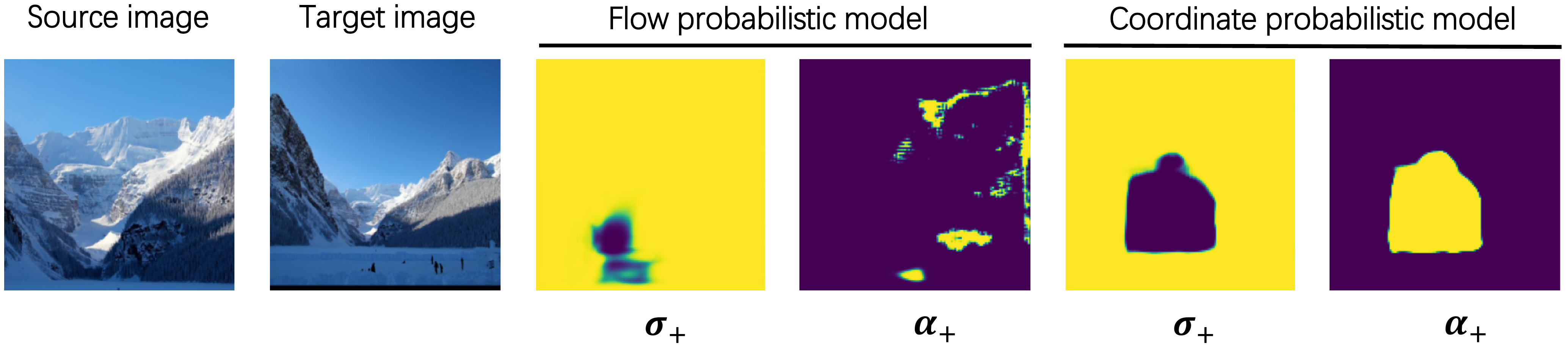}  \caption{\textbf{Parameters visualization of different model settings: flow probabilistic model \cite{truong2021learning} vs. coordinate probabilistic model (ours).} The parameters $\alpha_+$ and $\sigma_+$ of the target image are predicted with different probabilistic models. Our model works much better than flow probabilistic model.}
\label{fig:compare_model}
\end{figure}

\begin{figure}[!t]
\centering  
\includegraphics[width=0.48\textwidth]{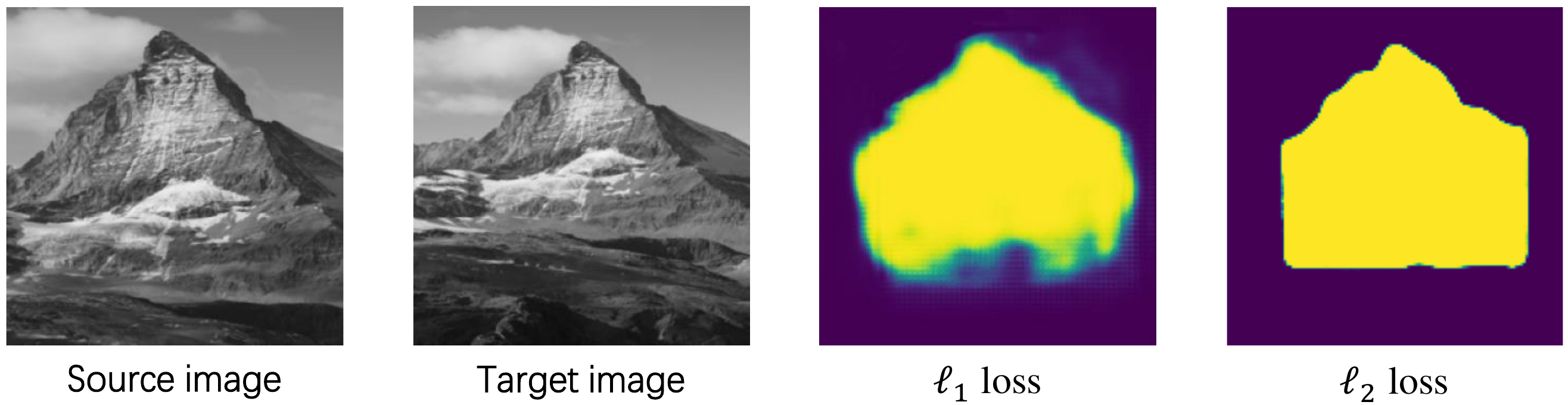}
\caption{\textbf{Visualization of confidence maps with the Laplacian model ($\ell_1$ loss) and the Gaussian model ($\ell_2$ loss).} The Gaussian model ($\ell_2$ loss) provides a sharper edge, which meets our requirement for the confidence map. } 
\label{fig:l1_vs_l2}
\end{figure}


To demonstrate the necessity of our probabilistic model, we compare our probabilistic coordinate model with the traditional probabilistic flow model, such as PDC-Net~\cite{truong2021learning}. The differences between parameterization are summarized: \romannumeral1) fixed $\sigma_+$ vs. learnable $\sigma_+$; \romannumeral2) the use of $\tt sigmoid$ instead of $\tt softmax$ to predict constrained $\alpha$; \romannumeral3) Laplacian model($\ell_1$ loss) vs. Gaussian model ($\ell_2$ loss). The reasons for our choice have been explained in Section~\ref{subsec:Coordinates in Conditional Modeling} and Section~\ref{subsec:Coordinates as Gaussian Mixture Models}. Here, we conduct an ablation experiment to identify their respective contributions.

\vspace{5pt}
\noindent\textbf{Experimental Details.} Based on the probabilistic flow model of PDC-Net~\cite{truong2021learning}, we replace the parameter settings of the original model in turn with our probabilistic coordinate model.
During the training stage, we only replaced the probabilistic model and the corresponding loss function in PCF-Net, and the other training settings remain the same. Note that here we only use one set of BCSs.  Table~\ref{table:pro_model} reports the Intersection-over-Union (IoU) metric between the union of different reliable regions and the ground truth flow map in two validation datasets (MegaDepth and ScanNet). The large performance gaps indicate that our coordinate probabilistic model is designed for this task. Specifically, substituting $\tt sigmoid$ for $\tt softmax$ to obtain the constrained $\alpha$ is crucial to help the entire network.
We visualize the prediction parameters $\sigma_+$ and $\alpha_+$, using the probabilistic model of PDCNet \cite{truong2021learning} that does not fit the distribution of coordinates directly as a baseline and our own model, respectively. As shown in Fig.~\ref{fig:compare_model}, the probabilistic model of PDCNet does not make sense, while our model yields reasonable prediction results. 

Meanwhile, we observe an interesting result, that the Gaussian model ($\ell_2$ loss) is more suitable for our task than the Laplacian model ($\ell_1$ loss) (discussed in Section~\ref{subsec:Coordinates in Conditional Modeling}). Here, we visualize their own predicted confidence maps in Fig.~\ref{fig:l1_vs_l2}. Clearly, the Gaussian model provides a sharper edge than the Laplacian model, which justifies our preference for the Gaussian model for probabilistic coordinate representations.

\begin{figure}[!t]
\centering  
\includegraphics[width=0.49\textwidth]{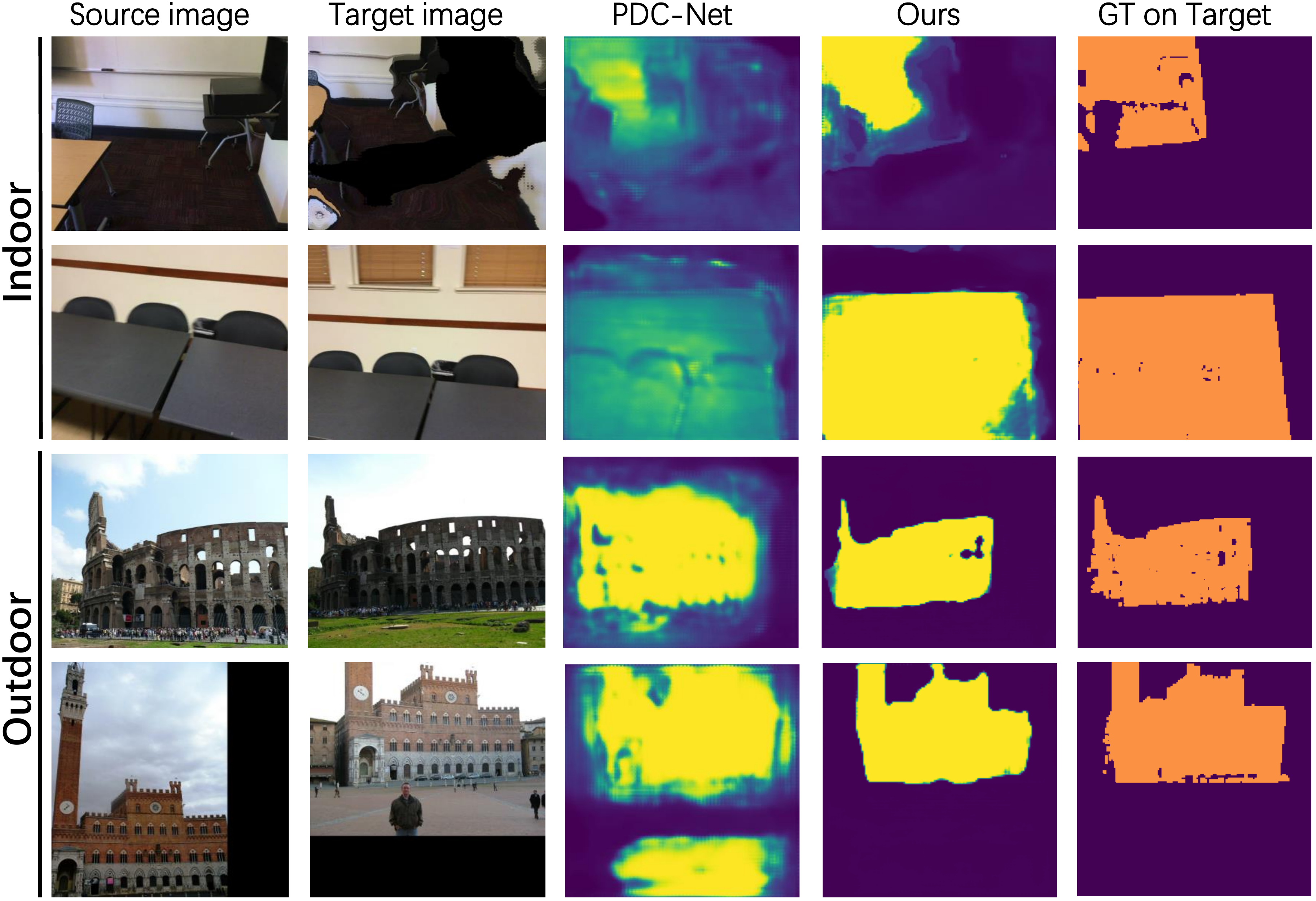}
\caption{\textbf{The accuracy of confidence maps for predicted flow fields.} In the $3^{rd}$ and $4^{th}$ columns, we visualize the confidence map of the target image using the probabilistic parameters predicted by PDC-Net~\cite{truong2021learning} and our PCF-Net, respectively. The brighter the regions, the more reliable they are. In the last column, we warp the ground truth flow into the target image and obtain the ground truth mask. Dark blue means that there is no matched correspondence between image pairs.} 
\label{fig:comparison}
\end{figure}

\begin{figure*}[!t]
    \centering
    \includegraphics[width=0.95\linewidth]{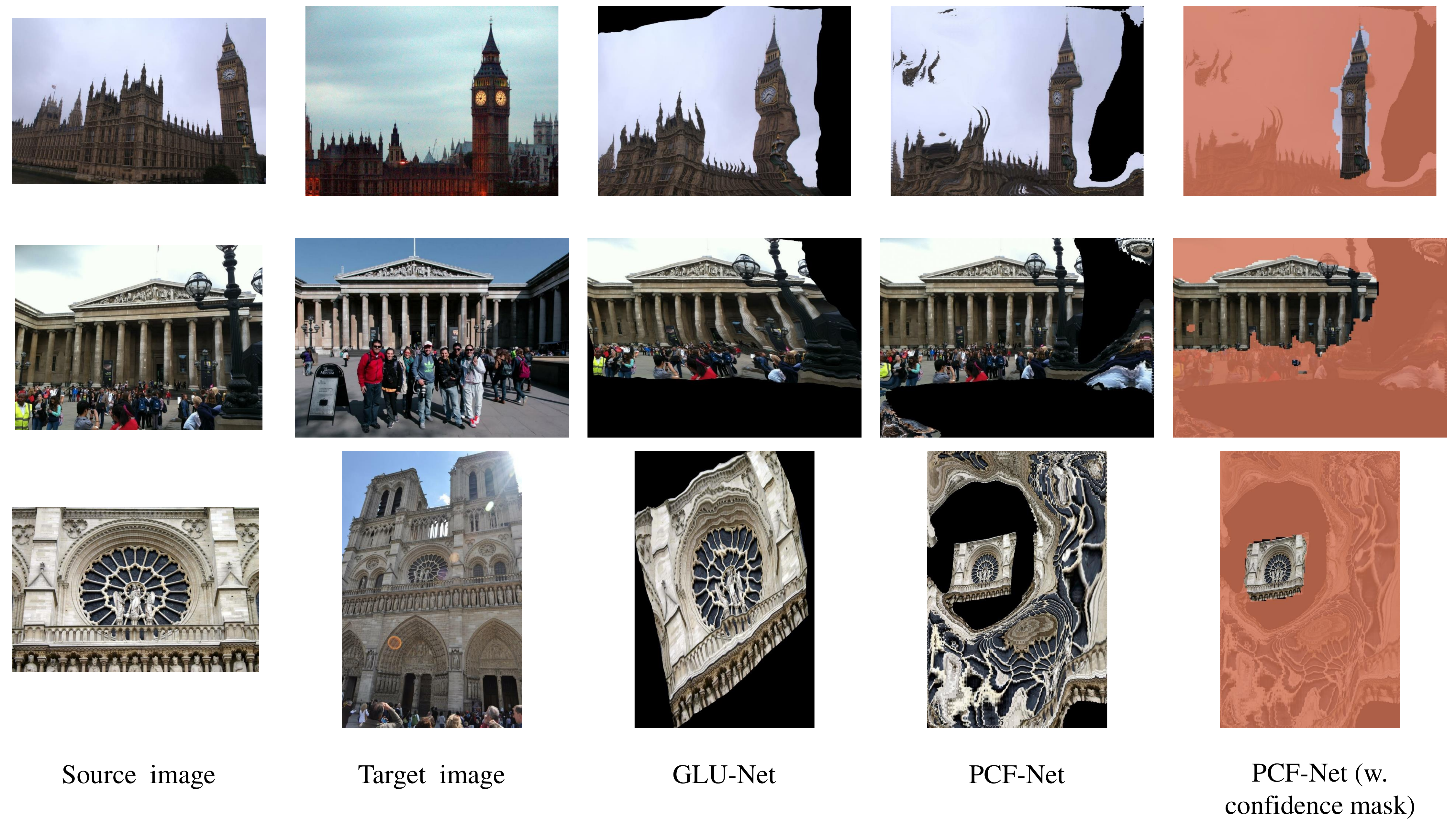}
    \caption{\textbf{Qualitative comparisons of our approach PCF-Net and GLU-Net, applied to images of the MegaDepth dataset. In the $3^{rd}$ and $4^{th}$ columns, we visualize the source images warped according to the flow estimated by the GLU-Net and PCF-Net respectively. In the lase column, PCF-Net predicts a confidence map, according to which the regions represented in red are unreliable.}}
    \label{pic:glu vs pcf}
\end{figure*}

\begin{figure*}[!t]
\centering  
\includegraphics[width=0.98\textwidth]{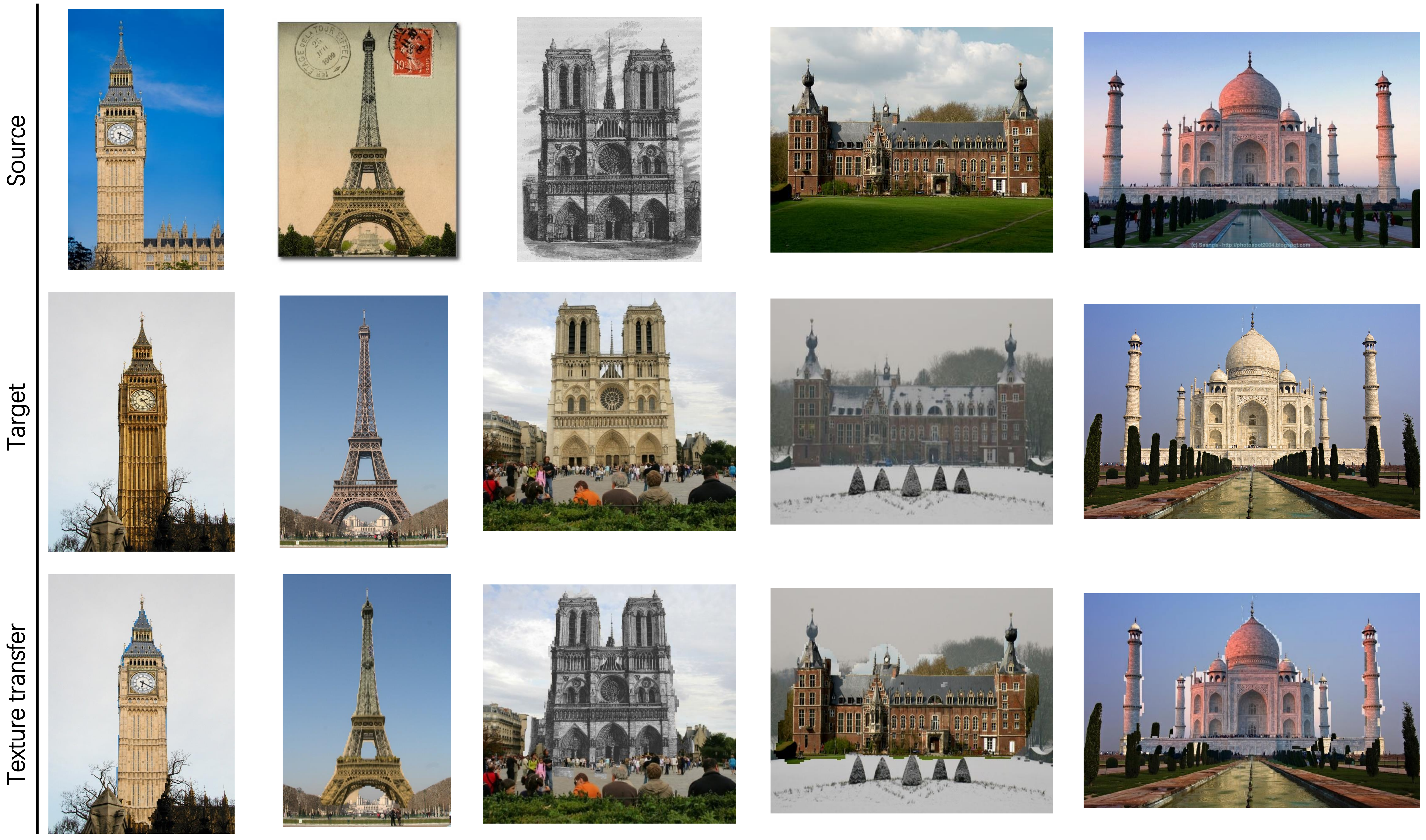}
\caption{\textbf{Texture transfer.} For the same architecture, we transfer textures from different photographs using the confidence map predicted by our approach, as well as the flow map. The contours of the architecture are perfectly divided, such as the Eiffel Tower and Big Ben. Note that we did not resort to any pre-trained segmentation models.} 
\label{fig:texture_transfer}
\end{figure*}

\subsection{Additional Analysis}

Our PCF-Net is built on the GLU-Net~\cite{truong2020glu}, which is a universal network architecture applicable to dense correspondence problems. To show the improvements brought about by our approach, we present a visual comparison of GLU-Net and our PCF-Net in Fig.~\ref{pic:glu vs pcf}. GLU-Net rarely considers the uncertainty of predicted correspondences, leading to large displacements and significant appearance transformations between image pairs (see the $3^{rd}$ column). Consequently, it cannot directly serve our probabilistic coordinate encoding task. By contrast, our PCF-Net integrates coordinate features and image features to estimate a confidence map, which can be used jointly for optical flow estimation and coordinate encoding. As illustrated in the last two columns of Fig.~\ref{pic:glu vs pcf}, our method achieves superior correspondence performance and more accurate confidence map estimation.

To further isolate the contribution of barycentric coordinates and reliable region prediction, we conduct the experiment on the MegaDepth dataset using LoFTR-OT~\cite{sun2021loftr} as a baseline. The position encoding module in LoFTR-OT will be replaced by the following three strategies: \romannumeral1) The PCF-Net without any modifications proposed in Section~\ref{subsec:Coordinates as Gaussian Mixture Models} and using cartesian coordinates (CC); \romannumeral2) The full proposed PCF-Net without distance map and using CC; \romannumeral3) The full proposed PCF-Net using the BCFs. It is important to note that each PCF-Net is trained individually based on the specific settings. The results are reported in Table~\ref{tab:contributions}, demonstrating that both the geometry-invariant barycentric coordinates and the proposed probabilistic models contribute to correspondences and pose estimation. It is worth noting that the original probability model can even lead to a decline in baseline performance, which highlights the necessity of the modifications proposed in our approach.

\begin{table}[!t] \scriptsize
\centering
\renewcommand{\arraystretch}{1.1}
\addtolength{\tabcolsep}{0.8pt}
\caption{Quantitative analysis of our contributions. CC: Cartesian coordinates without distance map illustrated in Section ~\ref{subsec:PCF-Net}.}

\begin{tabular}{@{}lccc@{}}
\toprule
Settings &AUC@5$^{\circ}$ &AUC@10$^{\circ}$ &AUC@20$^{\circ}$ \\ \midrule
 PCF-Net(w.o. modifications) + CC &45.64  &61.82  &75.37    \\
 PCF-Net(fully) + CC  &47.75 &63.93 &77.09   \\
 PCF-Net(fully) + BCFs   &\textbf{49.81} &\textbf{65.40} &\textbf{78.98}  \\
\bottomrule 
\end{tabular}
\label{tab:contributions}
\end{table}

\subsection{Extensions into Other Applications}

\noindent\textbf{Texture Transfer.} Since our PCF-Net method is built on the flow field, the confidence map for coordinate encoding can be used to represent the uncertainty of flow prediction. Mathematically, the coordinate confidence map is a subset of the flow confidence map. To better illustrate the relationship between them, we compare our confidence results with PDC-Net~\cite{truong2021learning}, the state-of-the-art network for predicting flow uncertainty.
When comparing the boundary differences between the predicted confidence maps and the GT, the prediction accuracy in the target contour varies between the approaches (as shown in Fig.~\ref{fig:comparison}).
Obviously, our approach generates a discriminative confidence map with a clear demarcation to classify regions. For a fair comparison, we train PDC-Net on the MegaDepth and ScanNet datasets following the original implementation, respectively. Interestingly, we emphasize the phenomenon that PDC-Net does not fit the probability distribution of indoor scenes well, while our approach works in different scenes.

Due to the potential of our approach to predict optical flow uncertainties and the sharp edges of approximate instance segmentation, our approach can be used to transfer texture between images. In Fig.~\ref{fig:texture_transfer} we show the results using historical and modern images from the LTLL dataset~\cite{LTLL}. Note that we did not resort to any pre-trained segmentation models. Perhaps our work may be useful for the task of segmentation.

\begin{figure}[!t]
\centering  
\includegraphics[width=0.40\textwidth]{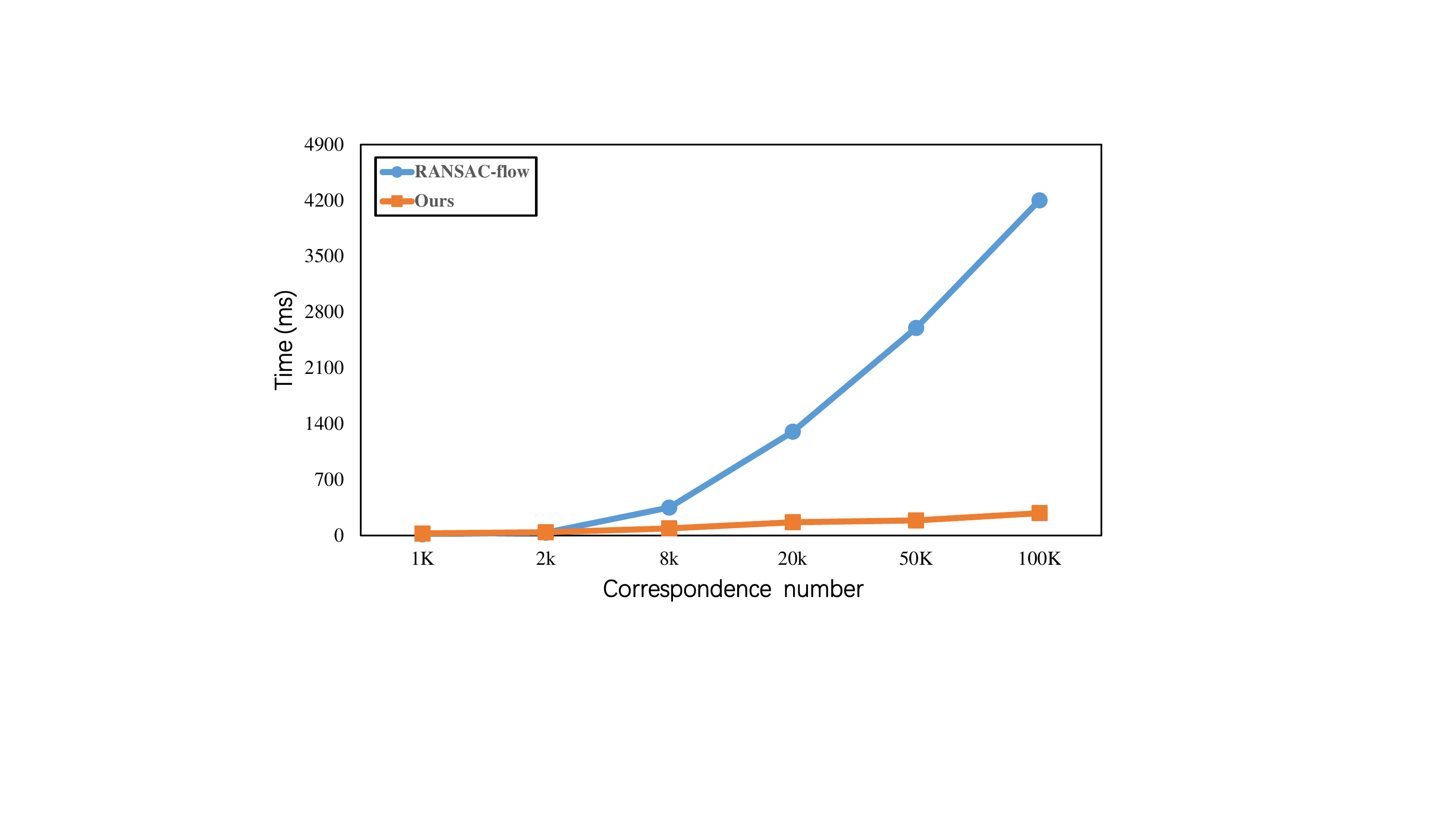}
\caption{\textbf{The efficiency of the computation compares the proposed multiple coordinate systems (Section~\ref{sec:strategies}) with RANSAC-flow~\cite{shen2020ransac}.} We take an image pair containing three homographies for testing. To control the correspondence number, we resize the image pair to different resolutions.} 
\label{fig:time}
\end{figure}

\vspace{5pt}
\noindent\textbf{Multi-homography Classification.}
Normally, we apply a single homography matrix to align the pair of images. However, such an assumption is invalid for image pairs with strong 3D effects or large object displacements. 
To relax this assumption, Multi-X~\cite{barath2018multi} formulated the multi-class model fitting by energy minimization and model searching. Recently, some work~\cite{AANAP,shen2020ransac} proposed iteratively reducing the threshold of RANSAC~\cite{ransac} to discover multiple homography candidates.
At each iteration, they can remove feature correspondences that are inliers for previous homographies (the threshold $\epsilon_g$) as well as from locations within the matchability masks predicted previously and recompute RANSAC using a smaller threshold ($\epsilon_l < \epsilon_g$ ). This procedure is repeated until the number of inliers is less than $\eta$. The image pairs are then divided into several nonoverlapping regions, each of which corresponds to a distinct homography transformation.

\begin{figure}[!t]
\centering  
\includegraphics[width=0.49\textwidth]{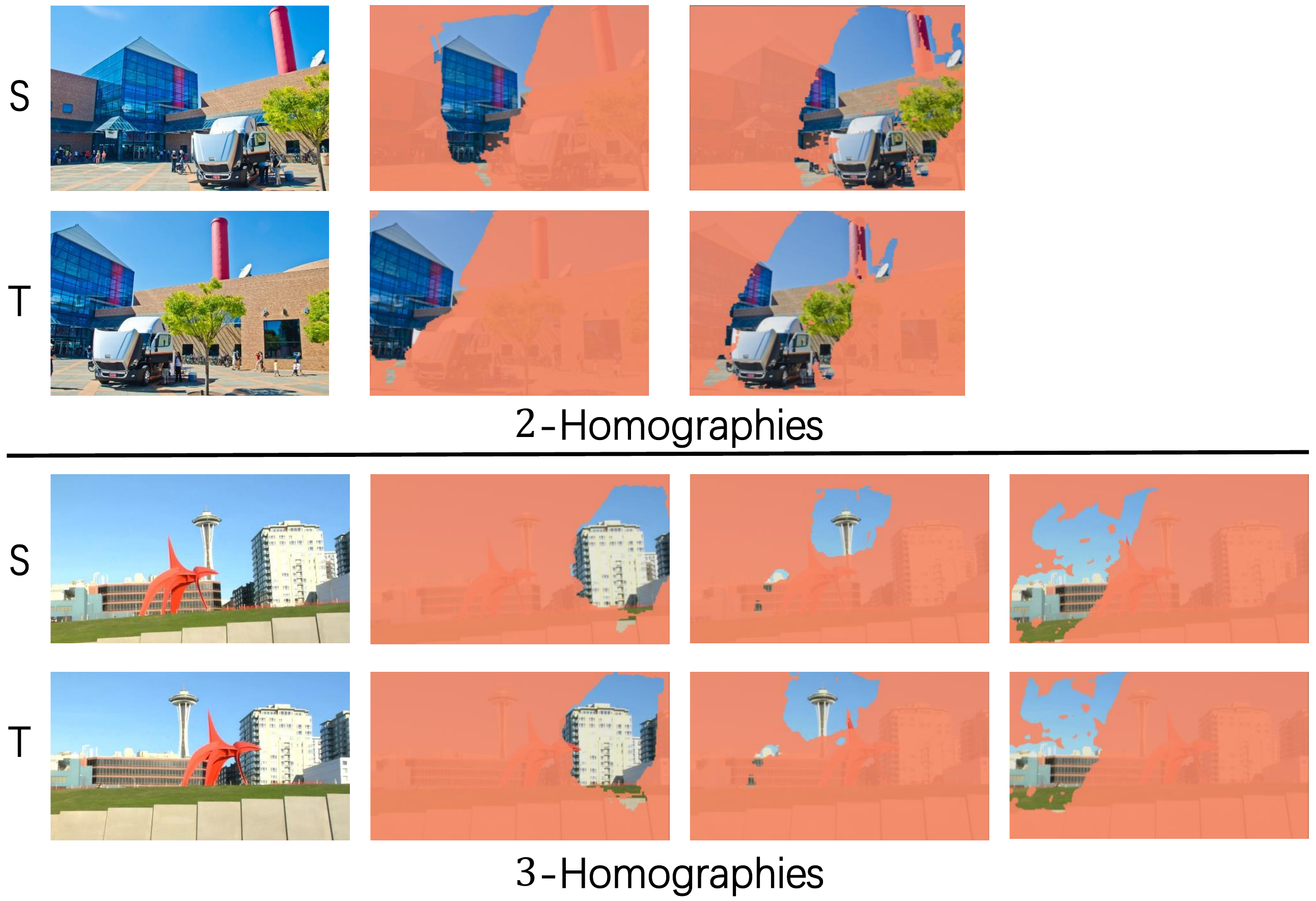}
\caption{\textbf{Multiple homography classification.} We visualize the $2$ homographies and $3$ homograhies examples from an image stitching dataset~\cite{zhang2014parallax}, respectively. Our approach can quickly and automatically divide them into multiple homography regions.} 
\label{fig:multi-affine}
\end{figure}

However, we found two drawbacks to such a strategy:  \romannumeral1) The selection of the threshold $\epsilon$ for each iteration is heuristic and sensitive for the results; \romannumeral2) When the number of input correspondences is large (\eg, dense flow map), the time consumption of the algorithm is intolerable, as shown in Fig.~\ref{fig:time}. For example, for an image pair of $720 \times 720$ resolution, the dense correspondences between them are at least $200$ K. 
Fortunately, our approach is not plagued by operational efficiency and hyperparameters. As proposed in Section~\ref{sec:strategies}, we simply need to change the mask area from a circle of fixed radius $(K-1)/2$ to the predicted reliable area. The number of iterations is pre-set like the RANSAC-flow~\cite{shen2020ransac}. We visualize the results in Fig.~\ref{fig:multi-affine}. For multiple homographies cases, our approach could successfully recognize them and give dense classification results. This may be helpful for image alignment or image stitching.

\subsection{Inference Time}
Our network and pipeline are implemented with PyTorch. Here we measure the inference time on an $24$ G NVIDIA GTX $3090$ GPU and Intel(R) Xeon(R) Gold $6226$R CPU @$2.90$GHz. The image pairs are of size $520 \times 520$, which corresponds to the predetermined input resolution of GLU-Net. Table~\ref{tab:runtime} is a summary of the run time of each component stage, including the extraction of feature maps, the estimation of flow, the calculation of BCFs, the prediction of probabilistic parameters and the calculation of PCFs.

Our pipeline takes $116$ ms on average to infer. For dense correspondence tasks~\cite{sun2021loftr,truong2020glu}, the extracted image feature maps can be shared with these methods to further save computational cost. The estimated flow map not only is used for our geometric-invariant coordinate encoding, but also contributes to other tasks, such as image alignment and texture transfer. The obtained confidence map can also be applied as the uncertainty of flow estimation, furthermore, to identify multi-homography regions. Overall, our proposed pipeline can be applied to different tasks with rapid inference time.

\begin{table}[!t]\footnotesize
    \centering
    \renewcommand{\arraystretch}{1.1}
    \addtolength{\tabcolsep}{1.2pt}
    \caption{Runtime of each component stage in our pipeline}
    \begin{tabular}{@{}lr@{}}
        \toprule
        Component stage  &Runtime\\ \midrule
        Feature maps extraction &18 ms\\
        Flow estimation &67 ms\\
        BCFs calculation &11 ms\\
        Probabilistic parameters prediction &15 ms\\
        PCFs calculation &5 ms\\ \midrule
        Total &116 ms\\
        \bottomrule
    \end{tabular}
\label{tab:runtime}
\end{table}

\begin{figure}[!t]
    \centering
    \subfigure[The comparison of time]{
    \includegraphics[width=0.40\textwidth]{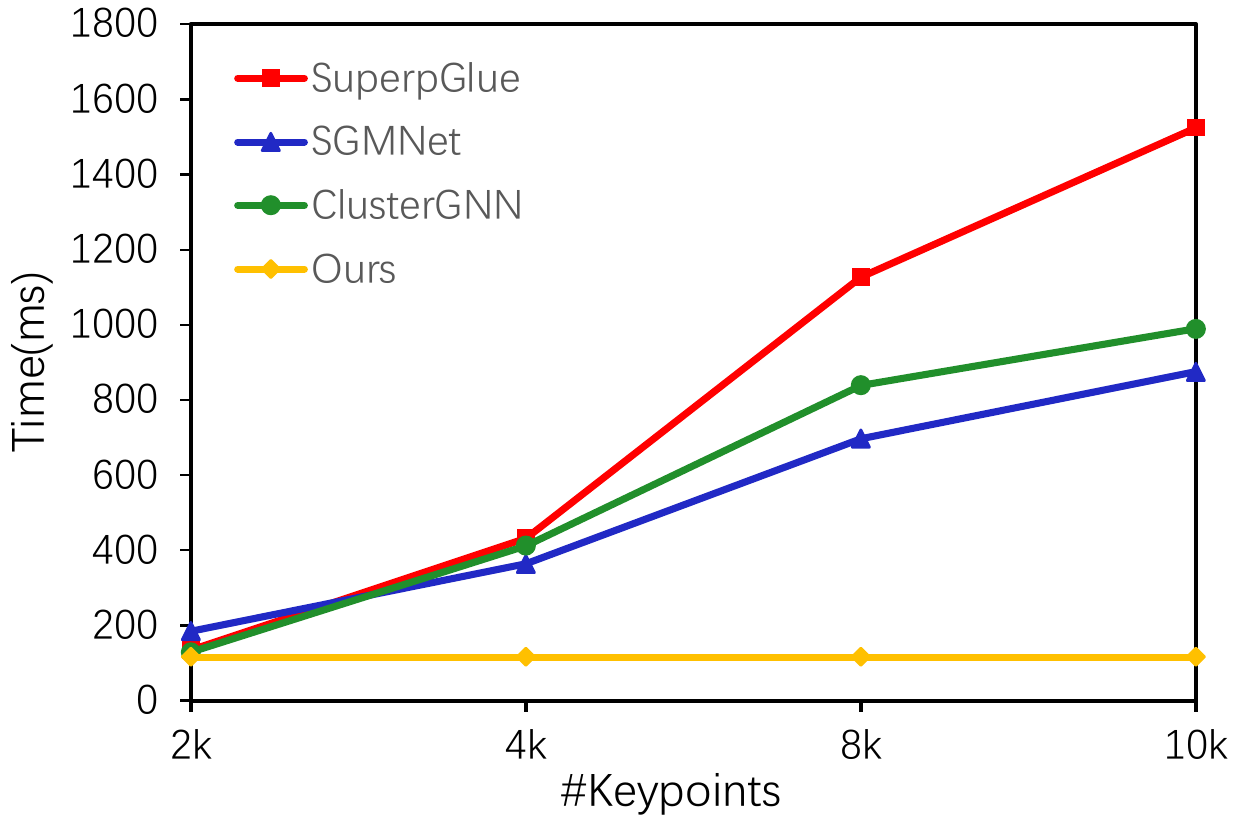}}
    \subfigure[The comparison of memory]{
    \includegraphics[width=0.40\textwidth]{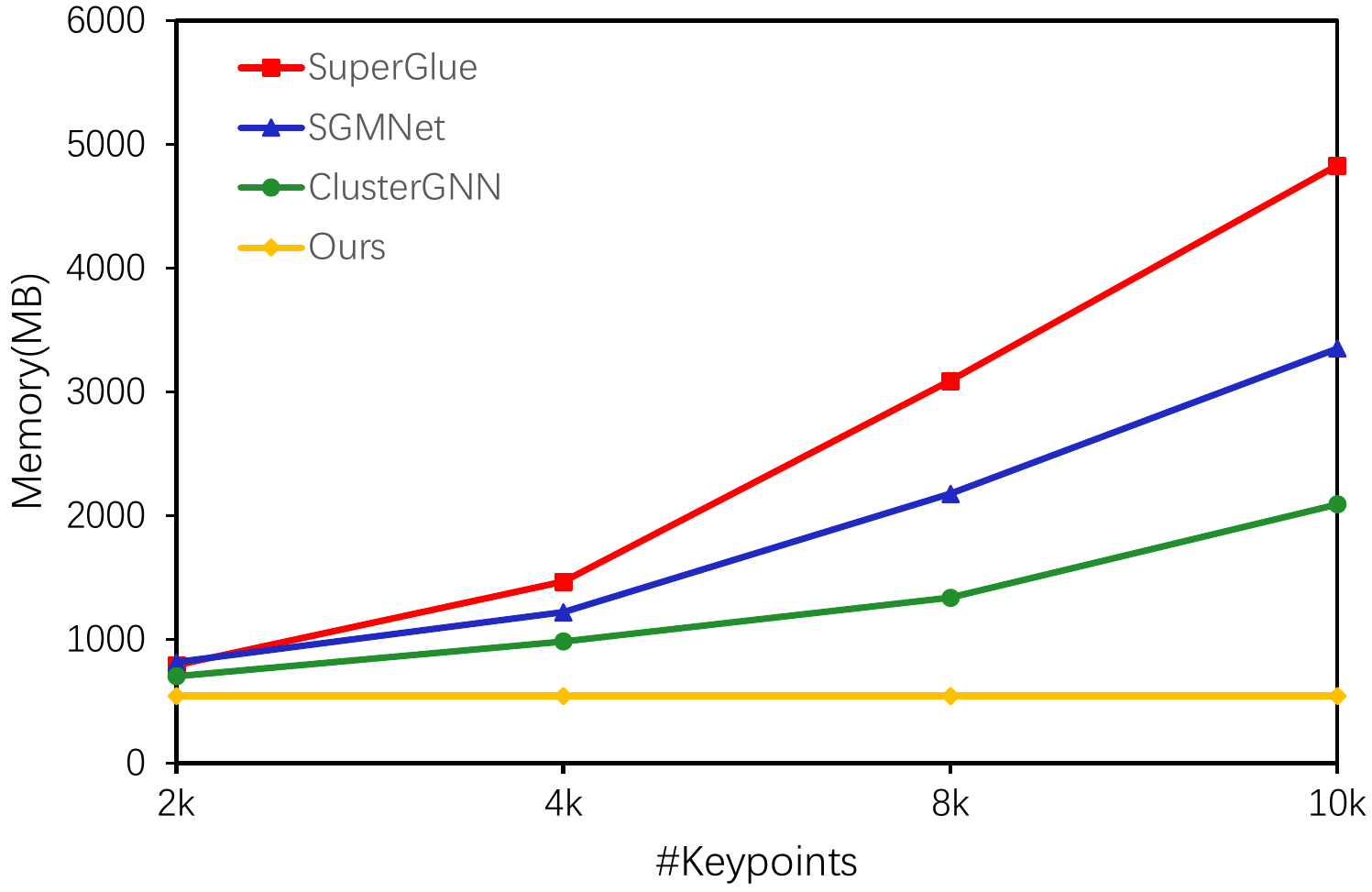}}
    \caption{Efficiency comparison. We report the time (a) and memory (b) consumption with increased number of input keypoints.}
\label{pic:comparsion efficiency}
\end{figure}

Furthermore, we also compared the PCF-Net with other state-of-the-art sparse matching methods, including SuperGlue, SGMNet~\cite{sgmnet} and ClusterGNN~\cite{clustergnn}. SGMNet and ClusterGNN both build on SuperGlue to further improve computational efficiency and matching performance. In particular, SGMNet establishes a small set of nodes to reduce the cost of attention. ClusterGNN uses K-means to construct local sub-graphs to save memory cost and computational overhead. We test the running time and memory of different methods with a gradually increasing number of keypoints. As shown in Fig.~\ref{pic:comparsion efficiency}(a), the runtime of sparse matching methods is relevant to the input number of keypoints. In particular, for the $10$K keypoints, the run-time of our method could be negligible compared to SuperGlue. Fig.~\ref{pic:comparsion efficiency}(b) reports memory occupation averaged by batch size.

\section{Conclusions}
In this work, we demonstrate the surprising performance of \textit{reliable} and \textit{geometric-invariant} coordinate representations for correspondence problems. Technically, we introduce the PCF and generate it using a PCF-Net network that jointly optimizes coordinate fields and confidence estimation. We show the effectiveness of PCF-Net in various problems and report highly consistent improved performance across multiple datasets. We believe that PCF-Net points out a novel direction for solving correspondence problems: learning reliable and geometric-invariant probabilistic coordinate representations. Future research directions include further optimization through cost aggregation~\cite{cho2022cats++} and graph matching~\cite{jiang2022robust}.

\bibliographystyle{IEEEtran}
\bibliography{ref}

\vfill

\begin{IEEEbiography}[{\includegraphics[width=1in,height=1.25in,clip,keepaspectratio]{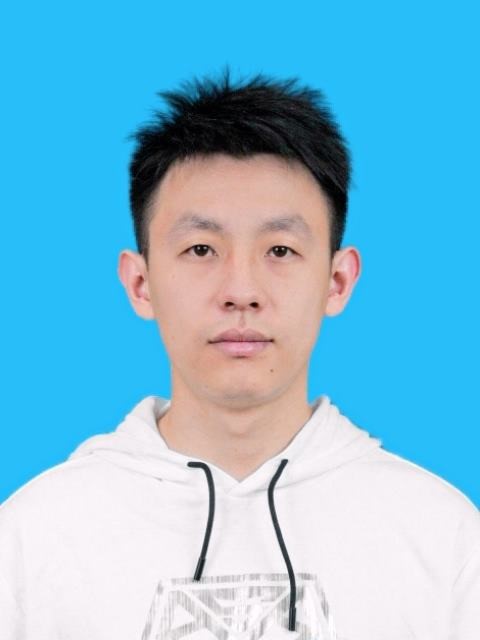}}]{Weiyue Zhao}
		 received the B.S. degree from Huazhong University of science and Technology, Wuhan, China, in 2020. He is currently pursuing the M.S. degree with the School of Artificial Intelligence and Automation, Huazhong University of Science and Technology, Wuhan, China.
 
 His research interests include computer vision and machine learning, with particular emphasis on image registration, multi-view stereo and various computer vision applications in video.
\end{IEEEbiography}

\vskip 0pt plus -1fil
\begin{IEEEbiography}[{\includegraphics[width=1in,height=1.25in,clip,keepaspectratio]{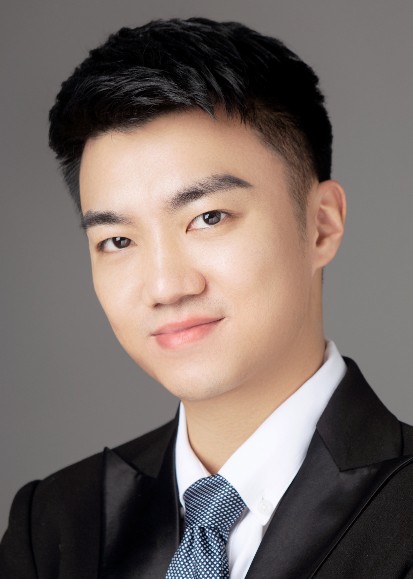}}]{Hao Lu}
		received the Ph.D. degree from Huazhong University of Science and Technology, Wuhan, China, in 2018.
  
  He was a Postdoctoral Fellow with the School of Computer Science, The University of Adelaide, Australia. He is currently an Associate Professor with the School of Artificial Intelligence and Automation, Huazhong University of Science and Technology, China. His research interest include various dense prediction problems in computer vision.
\end{IEEEbiography}

\vskip 0pt plus -1fil
\begin{IEEEbiography}[{\includegraphics[width=1in,height=1.25in,clip,keepaspectratio]{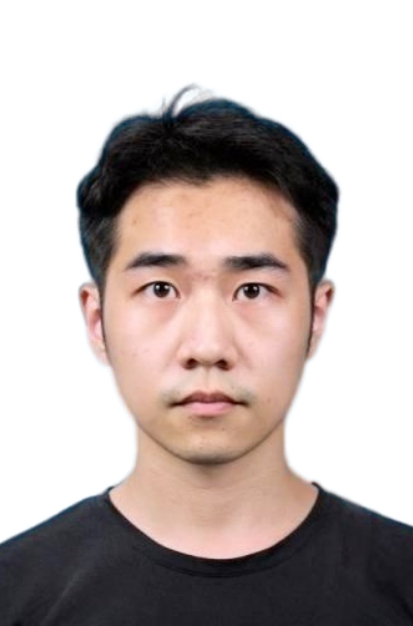}}]{Xinyi Ye}
		received the B.S. degree from Huazhong University of science and Technology, Wuhan, China, in 2021. He is currently pursuing the M.S. degree with the School of Artificial Intelligence and Automation, Huazhong University of Science and Technology, Wuhan, China.

His research interests include computer vision and machine learning, with particular emphasis on consistency filtering, multi-view stereo and inverse rendering for physics-based material editing and relighting.
\end{IEEEbiography}

\vskip 0pt plus -1fil
\begin{IEEEbiography}[{\includegraphics[width=1in,height=1.25in,clip,keepaspectratio]{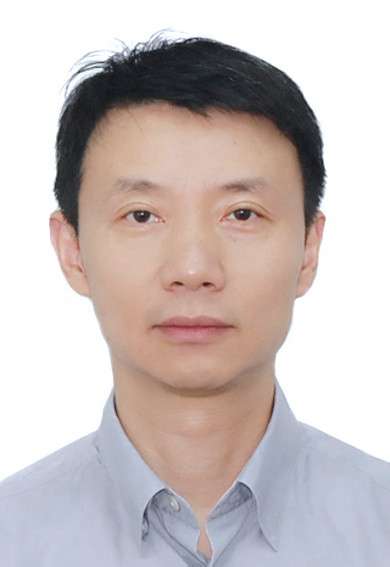}}]{Zhiguo Cao}
		received the B.S. and M.S. degrees in communication and information system from the University of Electronic Science and Technology of China, Chengdu, China, and the Ph.D. degree in pattern recognition and intelligent system from Huazhong University of Science and Technology, Wuhan, China.
  
  He is currently a Professor with the School of Artificial Intelligence and Automation, Huazhong University of Science and Technology, China. He has authored dozens of papers at international journals and conferences, which have been applied to automatic observation system for object recognition in video surveillance system, for crop growth in agriculture and for weather phenomenon in meteorology based on computer vision. His research interests spread across image understanding and analysis, depth information extraction, and object detection.
  
  Dr. Cao's projects have received provincial or ministerial level awards of Science and Technology Progress in China.
\end{IEEEbiography}

\vskip 0pt plus -1fil
\begin{IEEEbiography}[{\includegraphics[width=1in,height=1.25in,clip,keepaspectratio]{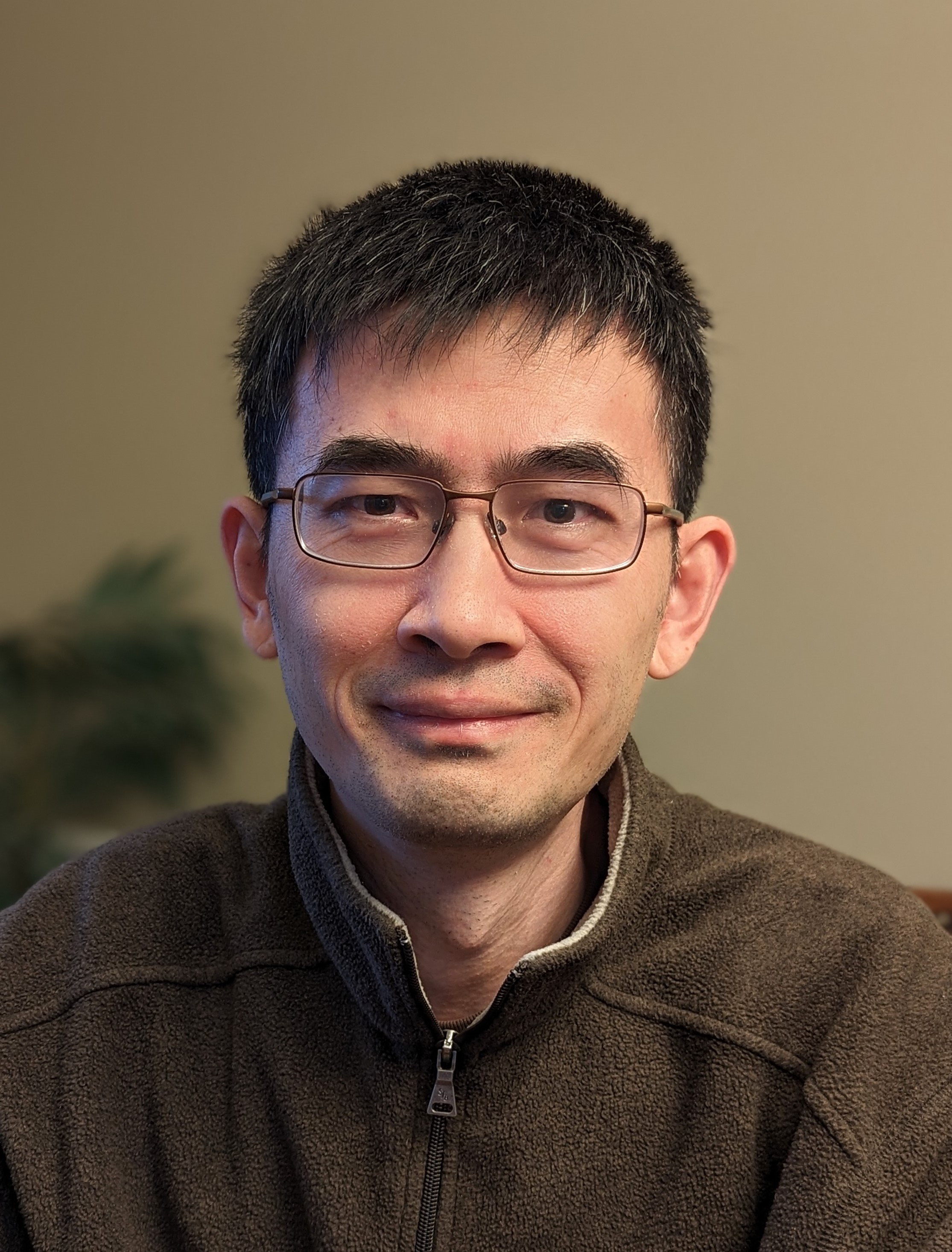}}]{Xin Li}
		received the B.S. degree with highest honors in electronic engineering and information science from University of Science and Technology of China, Hefei, in 1996, and the Ph.D. degree in electrical engineering from Princeton University, Princeton, NJ, in 2000. He was a Member of
Technical Staff with Sharp Laboratories of America, Camas, WA from Aug. 2000 to Dec. 2002. Since Jan. 2003, he has been a faculty member in Lane Department of Computer Science and Electrical Engineering. His research interests include image and video processing, compute vision and computational neuroscience. Dr. Li was elected a Fellow of IEEE in 2017 for his contributions to image interpolation, restoration and compression.

\end{IEEEbiography}	

\vfill

\ifCLASSOPTIONcaptionsoff
  \newpage
\fi

\end{document}